\documentclass{article}

\PassOptionsToPackage{numbers, compress}{natbib}

     \usepackage[preprint]{neurips_2021}

\usepackage[utf8]{inputenc} 
\usepackage[T1]{fontenc}    
\usepackage{hyperref}       
\usepackage{url}            
\usepackage{booktabs}       
\usepackage{amsfonts}       
\usepackage{nicefrac}       
\usepackage{microtype}      
\usepackage{xcolor}         

\usepackage{microtype}
\usepackage{graphicx}
\usepackage{subfigure}
\usepackage{booktabs} 

\usepackage{amsthm}
\usepackage{amssymb}
\usepackage{amsmath}

\DeclareMathOperator*{\argmin}{arg\,min}

\newtheorem{lem}{Lemma}
\newtheorem{pro}{Proposition}
\newtheorem{thm}{Theorem}
\newtheorem{corollary}{Corollary}

\newtheorem{assumption}{Assumption}

\usepackage{footnote}
\makesavenoteenv{tabular}
\usepackage{mathtools}
\usepackage{diagbox}
\usepackage{enumerate}

\usepackage{tablefootnote}

\allowdisplaybreaks[4]

\newtheorem{innercustomgeneric}{\customgenericname}
\providecommand{\customgenericname}{}
\newcommand{\newcustomtheorem}[2]{%
  \newenvironment{#1}[1]
  {%
   \renewcommand\customgenericname{#2}%
   \renewcommand\theinnercustomgeneric{##1}%
   \innercustomgeneric
  }
  {\endinnercustomgeneric}
}

\newcustomtheorem{lem_appendix}{Lemma}
\newcustomtheorem{pro_appendix}{Proposition}
\newcustomtheorem{thm_appendix}{Theorem}
\newcustomtheorem{corollary_appendix}{Corollary}
\newcustomtheorem{definition_appendix}{Definition}
\newcustomtheorem{assumption_appendix}{Assumption}

\title{Rethinking and Reweighting the Univariate Losses for Multi-Label Ranking: Consistency and Generalization}

\author{%
   Guoqiang Wu$^*$,~~ Chongxuan Li\thanks{Equal contribution} ,~~  Kun Xu,~~  Jun Zhu\thanks{Corresponding author} \\
   Dept. of Comp. Sci. \& Tech., Institute for AI, BNRist Center \\
   Tsinghua-Bosch Joint ML Center, THBI Lab, Tsinghua University, Beijing, 100084 China \\
  \texttt{\{guoqiangwu90, chongxuanli1991, kunxu.thu\}@gmail.com}, \\ \texttt{dcszj@mail.tsinghua.edu.cn}
}

\begin{document}

\maketitle

\begin{abstract}
(Partial) ranking loss is a commonly used evaluation measure for multi-label classification, which is usually optimized with convex surrogates for computational efficiency. Prior theoretical work on multi-label ranking mainly focuses on (Fisher) consistency analyses. However, there is a gap between existing theory and practice --- some pairwise losses can lead to promising performance but lack consistency, while some univariate losses are consistent but usually have no clear superiority in practice. In this paper, we attempt to fill this gap through a systematic study from two complementary perspectives of consistency and generalization error bounds of learning algorithms. Our results show that learning algorithms with the consistent univariate loss have an error bound of $O(c)$ ($c$ is the number of labels), while algorithms with the inconsistent pairwise loss depend on $O(\sqrt{c})$  as shown in prior work. This explains that the latter can achieve better performance than the former in practice. Moreover, we present an inconsistent reweighted univariate loss-based learning algorithm that enjoys an error bound of $O(\sqrt{c})$ for promising performance as well as the computational efficiency of univariate losses. Finally, experimental results validate our theoretical analyses.
\end{abstract}

\section{Introduction}
\label{sec:intro}
Multi-Label Classification (MLC)~\cite{mccallum1999multi} is an important task, in which each instance is associated with multiple labels simultaneously. It has a wide range of applications, such as text categorization~\cite{schapire2000boostexter}, bioinformatics~\cite{elisseeff2001kernel}, multimedia annotation~\cite{carneiro2007supervised}, and information retrieval~\cite{yu2005multi}. To evaluate the performance of different methods in MLC, various measures~\cite{zhang2013review, wu2017unified} have been developed from diverse aspects owing to the complexity of MLC. Among them, the \emph{(partial) ranking loss}~\cite{schapire2000boostexter, gao2013consistency}
is a widely-used measure in practice (or in theory).
Formally, the ranking loss calculates the fraction of pairs that a positive label does not precede a negative label according to the rank given by a \emph{score function} (or predictor). Accordingly, minimizing such a loss is usually referred to as Multi-Label Ranking (MLR)~\cite{dembczynski2012consistent}, which is our consideration in this paper. 

Since (partial) ranking loss is non-convex and discontinuous, existing methods~\cite{zhang2013review} seek to optimize certain convex surrogate losses for computational efficiency. These surrogate losses can be divided into two main categories: pairwise ones~\cite{gao2013consistency} and univariate ones~\cite{dembczynski2012consistent}, which have their own advantages and limitations in terms of computational costs, theory and empirical performance.

Computationally, the pairwise losses, defined over pairs of positive and negative labels, lead to a complexity depending on $O(c^2)$ ($c$ is the number of labels), while 
the univariate losses enjoy a  complexity depending on $O(c)$. Thus the latter are preferable, especially in cases with a large-scale label space.
Theoretically, the pairwise losses are not (Fisher) consistent w.r.t.  both the ranking loss and the partial ranking loss~\cite{gao2013consistency}, while, remarkably, certain univariate losses are consistent w.r.t.  the partial ranking loss~\cite{dembczynski2012consistent, gao2013consistency}. Empirically, however, the consistent univariate losses usually have no significant superiority in comparison with the inconsistent pairwise losses~\cite{dembczynski2012consistent}. In fact, we observed that the former under-perform  the latter on $10$  MLR benchmarks (see results in Table~\ref{tab:benchmark_results}). 
Such a gap between the existing theory and practice is worth being further studied and it would be appealing if one can further improve the performance of the univariate losses  especially when $c$ is large due to its computational efficiency.

\begin{table}[t]
\scriptsize
\centering
\caption{Summary of the main theoretical results. The contributions of this paper are highlighted in red.}
\label{main_theory_result}
\begin{minipage}{\textwidth}     
\begin{center}
\begin{small}
\begin{tabular}{ccccc}
\toprule
Algorithm & Surrogate loss & Generalization bound & Consistency\footnote{Note that this is in terms of partial ranking loss. Besides, these surrogate losses are all inconsistent w.r.t. ranking loss.} & Computational complexity \\
\midrule
$\mathcal{A}^{pa}$ & pairwise ($L_{pa}$) & $\hat{R}^{pa}_S(f) + O(\sqrt{\frac{c}{n}})$ & $\times$ & $O(c^2)$ \\
$\mathcal{A}^{u_1}$ & univariate ($L_{u_1}$) & $c \hat{R}^{u_1}_S(f) + O(\sqrt{\frac{c^2}{n}})$ & {\color{red}$\times$} & $O(c)$ \\
$\mathcal{A}^{u_2}$ & univariate ($L_{u_2}$) & {\color{red}$c \hat{R}^{u_2}_S(f) + O(\sqrt{\frac{c^2}{n}})$} & $\surd$\footnote{This is for the cases where the base loss is the exponential, logistic, least squared  or squared hinge loss.} & $O(c)$ \\
{\color{red}$\mathcal{A}^{u_3}$} & reweighted univariate ($L_{u_3}$)  & {\color{red}$\hat{R}^{u_3}_S(f) + O(\sqrt{\frac{c}{n}})$} & {\color{red}$\times$} & $O(c)$ \\
{\color{red}$\mathcal{A}^{u_4}$} & reweighted univariate ($L_{u_4}$) & {\color{red}$\hat{R}^{u_4}_S(f) + O(\sqrt{\frac{c^2}{n}})$} & {\color{red}$\times$ } & $O(c)$ \\
\bottomrule
\end{tabular}
\end{small}
\end{center}
\vskip -0.2in
\end{minipage}
\vskip -0.2in
\end{table} 

A natural explanation of the gap is that although the (Fisher) consistency~\cite{zhang2004statistical, bartlett2006convexity} provides valuable insights in the asymptotic cases, it cannot fully characterize the behaviour of a surrogate loss when the number of training samples is not sufficiently large and the hypothesis space is not realizable.

To address the issue, this paper presents a systematic study in a complementary perspective of generalization error bounds~\cite{mohri2018foundations} besides the consistency. In fact, we prove that the existing consistent univariate losses-based algorithms lead to an error bound depending on $O(c)$ while the pairwise losses based ones enjoy an error bound depending on $O(\sqrt{c})$~\cite{wu2020multi}, which explain the empirical behaviour better (see Table~\ref{tab:benchmark_results}). Further, 
we present two \emph{reweighted surrogate univariate losses} that employ carefully designed penalties for positive and negative labels. Such losses strictly upper bound the (partial) ranking loss, which is crucial in generalization analysis (see Section~\ref{sec:gene_bound}).
Moreover, we analyze their consistency and generalization bounds of the corresponding algorithms. 
Surprisingly, though not consistent, one of them enjoys an error bound depending on $O(\sqrt{c})$, which is nearly the same as the pairwise loss, and retains the computational efficiency. 
See Table~\ref{main_theory_result} for a summary of our main theoretical results.
Experimental results validate our theory findings.

Technically, focusing on  the widely used kernel-based algorithms~\cite{tan2020multi, wu2020joint,wu2020multi},  we present the generalization analyses based on Rademacher complexity~\cite{bartlett2002rademacher} and the vector-contraction inequality~\cite{maurer2016vector}, following recent work~\cite{wu2020multi}.
For the Fisher consistency, we consider more general reweighted univariate losses, which naturally extends the results in prior work~\cite{dembczynski2012consistent, gao2013consistency}. Considering different base losses (e.g., logistic loss), we present simple conditions that only involve penalties to characterize its consistency w.r.t.  (partial) ranking loss, which may be of independent interest.

This paper is organized as follows. In Section~\ref{sec:related_work}, we review the related work in MLC and MLR. Section~\ref{sec:preliminaries} introduces the problem setting, evaluation measures, risk and regret for MLR. Section~\ref{sec:methods} lists various surrogate losses, including the 
reweighted univariate ones, and their associated learning algorithms.
In section~\ref{sec:theoretical_analysis}, we present the generalization analyses of the algorithms and the consistency analyses of the corresponding surrogate losses.
Section~\ref{sec:experimental_results} presents and analyzes the experimental results. Section~\ref{sec:conclusion_future_work} concludes this paper and discusses future work.

\section{Related Work}
\label{sec:related_work}

Here we mainly review the theoretical work relevant to this paper in MLC and MLR.

\textbf{Consistency.} \cite{gao2013consistency} studied the consistency of various surrogate losses w.r.t. Hamming and (partial) ranking loss. Remarkably, \cite{dembczynski2012consistent} presented an explicit regret bound w.r.t. partial ranking loss for certain consistent univariate losses. Extensive work investigated the consistency w.r.t. other measures, especially the F-measure. For instance, \cite{ye2012optimizing} provided justifications and connections w.r.t. the F-measure using the empirical utility maximization (EUM) framework and the decision-theoretic approach (DTA) in binary classification, which were applied to the optimization of the macro-F measure in MLC. Further, \cite{dembczynski2017consistency} studied connections and differences between these two frameworks and clarified the notions of consistency\footnote{Note that, our generalization and consistency analyses are both under the EUM framework.} w.r.t. many complex measures (e.g., the F-measure and Jaccard measure) in binary classification. Besides, prior work~\cite{waegeman2014bayes,zhang2020convex} studied the consistency of the F-measure in MLC from the DTA perspective via different approaches to estimate the conditional distribution $P(\mathbf{y}|\mathbf{x})$. \cite{koyejo2015consistent} devoted to the study of consistent multi-label classifiers w.r.t. various measures under the EUM framework. \cite{menon2019multilabel} investigated the multi-label consistency of various reduction methods w.r.t. precision@$k$ and recall@$k$ measures.

\textbf{Generalization analysis.} \cite{wu2020multi} studied the generalization bounds of the algorithms based on the pairwise surrogate loss ($L_{pa}$) and the (variant)  univariate surrogate loss ($L_{u_1}$) w.r.t. the ranking loss. 

We mention that a specific form of Eq.~\eqref{eq:ral_surrogate_u3} with base hinge loss has been used as a part of prior work~\cite{xu2019robust}, which achieves excellent empirical results in MLC. In comparison, this paper considers a more general form of such reweighted surrogate losses and provides formal consistency and generalization analyses, which have not been investigated in the literature to our knowledge.

\section{Preliminaries}
\label{sec:preliminaries}

In this section, we fist introduce the problem setting of MLC and MLR. Then, we present the evaluation measures, risk, and regret of MLR.

\textbf{Notations}. Let boldface lower case letters denote vectors (e.g., $\mathbf{a}$) and boldface capital letters denote matrices (e.g., $\mathbf{A}$). 
For a matrix $\mathbf{A}$, $\mathbf{a}_i$, $\mathbf{a}^j$ and $a_{ij}$ denote its $i$-th row, $j$-th column, and $(i,j)$-th element respectively. 
For a vector $\mathbf{a}$, $a_{i}$ denote its $i$-th element.
For a square matrix, $\rm Tr(\cdot)$ denotes the trace operator. For a set, $|\cdot|$ denotes the cardinality. $[\![ \pi ]\!]$ denotes the indicator function, i.e., it returns $1$ when the proposition $\pi$ holds and $0$ otherwise. $sgn(x)$ returns $1$ when $x > 0$ and $-1$ otherwise.
$[n]$ denotes the set $\{1, ..., n\}$. For a function $g: \mathbb{R} \rightarrow \mathbb{R}$ and a matrix $\mathbf{A} \in \mathbb{R}^{m \times n}$, define $g(\mathbf{A}): \mathbb{R}^{m \times n} \rightarrow \mathbb{R}^{m \times n}$, where $g(\mathbf{A})_{ij} = g(a_{ij})$.

\subsection{Problem Setting}

Let $\mathbf{x} \in \mathcal{X} \subset \mathbb{R}^d$ and $\mathbf{y} \in \mathcal{Y} \subset \{ -1, +1 \}^c$ denote the input and output respectively, where $d$ is the feature dimension, $c$ is the number of labels, and the value $y_j = 1$ (or $-1$) indicates that the associated $j$-th label is relevant (or irrelevant). Given a training set $S = \{ ( \mathbf{x}_i, \mathbf{y}_i ) \}_{i=1}^n$ which is sampled i.i.d. from the distribution $P$ over $\mathcal{X} \times \mathcal{Y}$, the original goal of MLC is to learn a multi-label classifier $H: \mathbb{R}^d \longrightarrow \{ -1, +1 \}^c$. 

To solve MLC, a common approach is to first learn a vector-based \emph{score function} (or predictor) $f = [f_1, ..., f_c]: \mathbb{R}^d \longrightarrow \mathbb{R}^c$ and then get the classifier by a thresholding function. Multi-Label Ranking (MLR) aims to learn the best predictor from the finite training data in terms of some ranking-based measures, which is our consideration in this paper.

\subsection{Evaluation Measures}
To evaluate the performance of different approaches for MLR, many measures have been developed. Here we focus on two widely-used measures in practice (or theory), which are defined below\footnote{Our definition is over one sample and can be averaged over multiple samples.}.

\textbf{Ranking Loss}:
\begin{align}
    L_r^{0/1}(f(\mathbf{x}), \mathbf{y}) =  \frac{\sum_{(p, q) \in S_{\mathbf{y}}^+ \times S_{\mathbf{y}}^-}  [\![ f_p(\mathbf{x}) \leq f_q(\mathbf{x}) ]\!]}{|S_{\mathbf{y}}^+| |S_{\mathbf{y}}^-|}, \label{eqn:spe_rl}
\end{align}
where $S_{\mathbf{y}}^+$ (or $S_{\mathbf{y}}^-$) denotes the relevant (or irrelevant) label index set induced by $\mathbf{y}$.

\textbf{Partial Ranking Loss}\footnote{To minimize the partial ranking loss is equivalent to maximize the instance-AUC.}:
\begin{align}
    & L_{pr}^{0/1}(f(\mathbf{x}), \mathbf{y}) = \frac{1}{|S_{\mathbf{y}}^+| |S_{\mathbf{y}}^-|} \sum_{(p, q) \in S_{\mathbf{y}}^+ \times S_{\mathbf{y}}^-} \bigg[ [\![ f_p(\mathbf{x}) < f_q(\mathbf{x}) ]\!]
     + \frac{1}{2} [\![ f_p(\mathbf{x}) = f_q(\mathbf{x}) ]\!] \bigg  ]. \label{eqn:spe_prl}
\end{align}

From the above definitions, we can observe that the only difference between these two measures is the penalty when $f_p(\mathbf{x}) = f_q(\mathbf{x})$ holds. Besides, it is easy to verify that ranking loss upper bounds the partial ranking loss, i.e. $L_{pr}^{0/1}(f(\mathbf{x}), \mathbf{y}) \leq L_r^{0/1}(f(\mathbf{x}), \mathbf{y})$. Although these two measures are almost the same in practice for the evaluation of one algorithm, they have different consistency properties for some surrogate losses theoretically~\cite{gao2013consistency}.

\subsection{Risk and Regret}

Since (partial) ranking loss is non-convex and discontinuous, often leading to NP-hard problems \cite{arora2009computational}, extensive methods optimize it with convex surrogate losses in practice for computational efficiency. Define a surrogate loss $L_{\phi}: \mathbb{R}^c \times \{-1, +1\}^c \rightarrow \mathbb{R}_+$, where $\phi$ indicates the specific surrogate loss and will be detailed in the next section. Besides, define a vector-based predictor class $\mathcal{F} = \{ f: \mathcal{X} \mapsto \mathbb{R}^c \}$. For a predictor $f \in \mathcal{F}$, its true ($0/1$) expected risk, surrogate expected risk, and surrogate empirical risk are defined as follows:
\begin{align}
    R_{0/1}(f) & = \mathop{\mathbb{E}}_{(\mathbf{x}, \mathbf{y}) \sim P} [ L^{0/1} (f(\mathbf{x}), \mathbf{y}) ], \\
 R_{\phi}(f) & = \mathop{\mathbb{E}}_{(\mathbf{x}, \mathbf{y}) \sim P} [ L_{\phi} (f(\mathbf{x}), \mathbf{y}) ], \\ \hat{R}_{S}(f) & = \frac{1}{n} \sum_{i=1}^n  L_{\phi} (f(\mathbf{x}_i), \mathbf{y}_i).
\end{align}
Besides, we use a superscript (i.e., $pr$ or $r$) to distinguish the risks for specific measures. For instance, $\hat{R}_{S}^{pr}(f)$ and $\hat{R}_{S}^{r}(f)$ denote the empirical partial ranking risk and the empirical ranking risk respectively. Moreover, for convenience the expected risk conditioned on an instance $\mathbf{x}$ (i.e., the conditional risk) can be expressed as:
\begin{align}
    R(f | \mathbf{x}) & = \mathop{\mathbb{E}}_{\mathbf{y} \sim P(\mathbf{y} | \mathbf{x})} [ L (f(\mathbf{x}), \mathbf{y}) | \mathbf{x}] 
    = \sum_{\mathbf{y}}  L (f(\mathbf{x}), \mathbf{y}) P(\mathbf{y} | \mathbf{x}),
\end{align}
where $L$ denotes the true ($0/1$) or surrogate loss. Thus, the expected risk of $f$ is $R(f) = \mathop{\mathbb{E}}_{\mathbf{x} \sim P(\mathbf{x})} [R(f | \mathbf{x})]$.

For each $\mathbf{x}$, given the conditional distribution $P(\mathbf{y} | \mathbf{x})$, we can get its optimal predictions\footnote{Notably, the optimal predictions can be not just one value but a set with many elements that share the same minimal conditional risk.} as follows.
\begin{align}
    f^*(\mathbf{x}) = \argmin_{\mathbf{a} \in \mathbb{R}^c} \sum_{\mathbf{y}}  L (\mathbf{a}, \mathbf{y}) P(\mathbf{y} | \mathbf{x}),
\end{align}
where $f^*$ is called the \emph{Bayes predictor} w.r.t. the loss $L$. Besides, the expected risk of $f^*$ (i.e., $R(f^*)$) is called the \emph{Bayes risk}, which is the minimal expected risk w.r.t. the loss $L$ and denoted by $R^*$ for convenience. Then, we can define the regret (a.k.a. excess risk) of a  predictor  $f$ w.r.t. the true and surrogate loss as follows.
\begin{align}
    Reg_{0/1}(f) & = R_{0/1}(f) - R_{0/1}^*, \\
    Reg_{\phi}(f) & = R_{\phi}(f) - R_{\phi}^*.
\end{align}
Besides, we also use a superscript (i.e., $pr$ or $r$) to distinguish the regrets for specific measures. Moreover, we denote the learned  predictor from finite training data $S$ as $\hat{f}_n$. Note that, our goal is to find a predictor $\hat{f}_n$ that achieves the minimal true regret (i.e. $Reg_{0/1}(\hat{f}_n)$) as possible as it can.

\section{Methods}
\label{sec:methods}

In this section, we first introduce several specific surrogate losses. Then, we present their associated learning algorithms.

\subsection{Surrogate losses}
\label{sec:loss}

To optimize the (partial) ranking loss, it is natural to employ the convex surrogate pairwise loss~\cite{schapire2000boostexter, elisseeff2001kernel, dekel2003log, zhang2006multilabel} as follows:
	\begin{equation}
	\label{eq:ral_surrogate_p}
		L_{pa} (f(\mathbf{x}), \mathbf{y}) = \frac{1}{{|S_{\mathbf{y}}^+| |S_{\mathbf{y}}^-|}} \sum_{(p, q) \in S_{\mathbf{y}}^+ \times S_{\mathbf{y}}^-}  \ell ( f_p(\mathbf{x}) - f_q(\mathbf{x}) ) .
	\end{equation}
where the base (margin-based) convex loss $\ell(z)$ can be defined in various popular forms, such as the exponential loss $\ell(z) = e^{-z}$, the logistic loss $\ell(z) = \ln( 1 + e^{-z} )$, the hinge loss $\ell(z) = \max \{0, 1 - z\}$, and squared hinge loss $\ell(z) = (\max \{0, 1 - z\})^2$. A common property is that the base convex surrogate loss upper bounds the original $0/1$ loss\footnote{The original logistic loss can be easily changed to $\ell(z) = \log_2( 1 + 2^{-z} )$ or $\ell(z) = \ln( e - 1 + e^{-z} )$ to satisfy this condition.}, i.e., $[\![ z \leq 0 ]\!] \leq \ell(z)$.

Besides, the surrogate univariate loss, which primarily aims to optimize Hamming loss~\cite{boutell2004learning, wu2020multi}, can also be viewed as a surrogate loss for the (partial) ranking loss, which is defined as follows:
	\begin{equation}
	\label{eq:ral_surrogate_u1}
		L_{u_1} (f(\mathbf{x}), \mathbf{y}) = \frac{1}{c}
		\sum_{j=1}^c \ell(y_{j} f_j(\mathbf{x})).
	\end{equation}
Note that $L_{u_1}$ cannot strictly upper bound the (partial) ranking loss, i.e. $L_r^{0/1}(f(\mathbf{x}), \mathbf{y}) \nleq L_{u_1} (f(\mathbf{x}), \mathbf{y})$.

Remarkably, previous work presents the consistent surrogate univariate loss~\cite{dembczynski2012consistent, gao2013consistency} w.r.t. partial ranking loss, which is defined as follows:
	\begin{equation}
	\label{eq:ral_surrogate_u2}
		L_{u_2} (f(\mathbf{x}), \mathbf{y}) = \frac{1}{|S_{\mathbf{y}}^+| |S_{\mathbf{y}}^-|}
		\sum_{j=1}^c \ell(y_{j} f_j(\mathbf{x})).
	\end{equation}
Again, the consistent surrogate loss $L_{u_2}$ cannot strictly upper bound the (partial) ranking loss either. Notably, when the surrogate loss strictly upper bounds the $0/1$ loss, the true ($0/1$) risk can be upper bounded by the surrogate risk too, which is crucial for its generalization analysis.  Thus, we present two \emph{reweighted convex surrogate univariate losses}, which strictly upper bound (partial) ranking loss, defined as below. 
	\begin{equation}
	\label{eq:ral_surrogate_u3}
		L_{u_3} (f(\mathbf{x}), \mathbf{y}) = \frac{\sum_{p \in S_{\mathbf{y}}^+} \ell(y_{p} f_p(\mathbf{x}))}{|S_{\mathbf{y}}^+|} + \frac{\sum_{q \in S_{\mathbf{y}}^-} \ell(y_{q} f_q(\mathbf{x}))}{|S_{\mathbf{y}}^-|},
	\end{equation}
	\begin{equation}
	\label{eq:ral_surrogate_u4}
		L_{u_4} (f(\mathbf{x}), \mathbf{y}) = \frac{1}{\min\{|S_{\mathbf{y}}^+|, |S_{\mathbf{y}}^-|\}}
		\sum_{j=1}^c \ell(y_{j} f_j(\mathbf{x})).
	\end{equation}
For a clear presentation, we will formally discuss the  relationships among these surrogate losses in the next section.

\subsection{Learning Algorithms}
In the following, we consider the kernel-based learning algorithms which have been widely used in practice~\cite{elisseeff2001kernel, boutell2004learning, hariharan2010large, tan2020multi, wu2020joint} and in theory~\cite{wu2020multi} in MLC. Besides, our following analyses can be extended to other forms of hypothesis class, such as neural networks~\cite{anthony2009neural}. Let $\mathbb{H}$ be a reproducing kernel Hilbert space (RKHS) induced by the kernel function $\kappa$, where $\kappa: \mathcal{X} \times \mathcal{X} \rightarrow \mathbb{R}$ is a Positive Definite Symmetric (PSD) kernel. Let $\Phi: \mathcal{X} \rightarrow \mathbb{H}$ be a feature mapping associated with $\kappa$. The kernel-based hypothesis class can be defined as follows.
    \begin{align}
    \label{eq:kernel_hypothesis}
        \mathcal{F} = \bigg\{ \mathbf{x} \mapsto \mathbf{W} ^\top \phi(\mathbf{x}): \mathbf{W}= (\mathbf{w}_1, \ldots ,\mathbf{w}_c)^\top, \| \mathbf{W} \| \leq \Lambda \bigg\} ,
    \end{align}
where $\| \mathbf{W} \|$ denotes $\| \mathbf{W} \|_{\mathbb{H}, 2} = (\sum_{j=1}^c \| \mathbf{w}_j \|_{\mathbb{H}}^2)^{1/2}$ for convenience.

Here we consider the following five learning algorithms with the corresponding aforementioned surrogate losses.
    \begin{align}
        \label{alg:pa}
		& \mathcal{A}^{pa}: \ \min_{\mathbf{W}} \ \frac{1}{n} \sum_{i=1}^n L_{pa}(f(\mathbf{x}_i), \mathbf{y}_i) + \lambda \| \mathbf{W} \|^2 ,\\
        \label{alg:u}
		& \mathcal{A}^{u_k}: \ \min_{\mathbf{W}} \ \frac{1}{n} \sum_{i=1}^n L_{u_k}(f(\mathbf{x}_i), \mathbf{y}_i) + \lambda \| \mathbf{W} \|^2 , k = 1, 2, 3, 4.
    \end{align}

\section{Theoretical Analyses}
\label{sec:theoretical_analysis}
In this section, we present generalization error bounds of the learning algorithms presented before and consistency analyses of the corresponding surrogate losses.

Firstly, we want to highlight the complementary roles of the two perspectives. Recall that our goal is to find a predictor $\hat{f}_n$ learned from finite training data that achieves the minimal true regret $Reg_{0/1}(\hat{f}_n)$. In the following, we will decompose the regret appropriately for clear discussions.  

For generalization analyses, the true regret can be decomposed into the following terms w.r.t. the $0/1$ loss. 
\begin{equation}
    \begin{split}
        & Reg_{0/1}(\hat{f}_n) = R_{0/1} (\hat{f}_n) - R_{0/1}^*  = \underbrace{\left[ R_{0/1} (\hat{f}_n) - \inf_{g \in \mathcal{F}} R_{0/1} (g) \right]}_{\text{estimation error}} + \underbrace{\left[ \inf_{g \in \mathcal{F}} R_{0/1} (g) - R_{0/1}^* \right]}_{\text{approximation error}},
    \end{split}
\end{equation}
where $\mathcal{F}$ is the constrained function class that real learning algorithms utilize. For a given distribution $P(\mathbf{x}, \mathbf{y})$ and a specific measure, $R_{0/1}^*$ is fixed. Besides, $\inf_{g \in \mathcal{F}} R_{0/1} (g)$ depends on the size of $\mathcal{F}$ and is fixed for a given $\mathcal{F}$. Thus, in this case, the original goal becomes to minimize $R_{0/1} (\hat{f}_n)$ as possible as it can. In Section~\ref{sec:gene_bound}, we present the generalization error bounds of the learning algorithms to provide learning guarantees for $R_{0/1} (\hat{f}_n)$ through bounding the surrogate risk $R_{\phi} (\hat{f}_n)$\footnote{Note that, this requires that the surrogate loss $L_{\phi}$ strictly upper bounds the $0/1$ loss $L_{0/1}$ to make $R_{0/1} \leq R_{\phi}$.}. However, these error bounds cannot exactly tell the size of the gap between $R_{0/1} (\hat{f}_n)$ and $R_{\phi} (\hat{f}_n)$.

Consistency analyses aim to answer the question whether the ($0/1$) expected risk of the learned function converges to the Bayes risk~\cite{bartlett2006convexity,gao2013consistency}, i.e., when $n \rightarrow \infty$, $R_{\phi} (\hat{f}_n) \rightarrow R_{\phi}^* \Longrightarrow R_{0/1} (\hat{f}_n) \rightarrow R_{0/1}^*$. If a loss is consistent, a regret bound~\cite{bartlett2006convexity, dembczynski2012consistent} as follows is preferable. Namely, for all measurable function $f$ (including $\hat{f}_n$) and valid joint distribution $P(\mathbf{x}, \mathbf{y})$, the following holds:
\begin{align} 
R_{0/1} (f) - R_{0/1}^* \leq \psi^{-1} (R_{\phi} (f) - R_{\phi}^*),
\end{align}
where $\psi$ is an invertible function such that for any sequence $(\theta_i)$ in $[0,1]$, $\psi(\theta_i) \rightarrow 0$ if and only if $\theta_i \rightarrow 0$~\cite{bartlett2006convexity}.
Prior work~\cite{dembczynski2012consistent} shows that $\psi^{-1}(\theta) = O(c\sqrt{c}) \sqrt{\theta}$ with logistic and exponential loss in MLR.
Besides, when learning in the real setting (with finite data), the surrogate regret of $\hat{f}_n$ can be decomposed into the following two terms w.r.t. the surrogate loss.
\begin{equation}
    \begin{split}
        & \quad R_{\phi} (\hat{f}_n) - R_{\phi}^* = \underbrace{\left[ R_{\phi} (\hat{f}_n) - \inf_{g \in \mathcal{F}} R_{\phi} (g) \right]}_{\text{estimation error}} + \underbrace{\left[ \inf_{g \in \mathcal{F}} R_{\phi} (g) - R_{\phi}^* \right]}_{\text{approximation error}},
    \end{split}
\end{equation}
where the estimation error is due to finite data size, and the approximation error is due to the choice of $\mathcal{F}$.
Notably, the consistency analysis~\cite{bartlett2006convexity} neglects these two errors since it allows $P(\mathbf{y} | \mathbf{x})$ known in the infinite data setting and assumes that the hypothesis class $\mathcal{F}$ is over all measurable functions.

In summary, consistency can provide valuable insights for learning from infinite data (or data of relatively large $n$ w.r.t. $c$) with an unconstrained hypothesis class, while generalization bounds can offer more insights for learning from finite data with a constrained hypothesis class.

\subsection{Generalization Analyses}
\label{sec:gene_bound}
For generalization analyses, we mainly follow the recent theoretical work \cite{wu2020multi}. 
First, we introduce the common assumptions for the subsequent analyses.
    \begin{assumption}[The common assumptions]
    \label{assump_1}
    $ $
        \begin{enumerate}[(1)]
            \item The hypothesis class is defined in Eq.\eqref{eq:kernel_hypothesis}.
            \item The training dataset $S = \{ ( \mathbf{x}_i, \mathbf{y}_i ) \}_{i=1}^n$ is sampled i.i.d. from the distribution $P$, where $\exists \ r > 0$, it satisfies $\kappa (\mathbf{x}, \mathbf{x}) \leq r^2$ for all $\mathbf{x} \in \mathcal{X}$.
		    \item The base (convex) loss $\ell (z)$ is $\rho$-Lipschitz continuous and bounded by $B$. 
        \end{enumerate}
    \end{assumption}
    Note that, the widely-used hinge and logistic loss are both $1$-Lipschitz continuous\footnote{Although the exponential, and squared hinge losses are not globally Lipschitz continuous, they are locally Lipschitz continuous.}. Then we provide the properties of surrogate losses in the following lemma. Notably, the Lipschitz constants of surrogate losses characterize the relationship between the Rademacher complexities~\cite{bartlett2002rademacher} of the loss class and hypothesis class based on the vector-contraction inequality~\cite{maurer2016vector}, which plays a central role in the generalization analysis.
    \begin{lem}[The properties of surrogate losses; full proof in Appendix A.1]
        Assume that the base (convex) loss $\ell (z)$ is $\rho$-Lipschitz continuous and bounded by $B$. Then, the following holds. 
        \begin{enumerate}[(1)]
            \item the surrogate loss $L_{u_2} (f(\mathbf{x}), \mathbf{y})$ in Eq.\eqref{eq:ral_surrogate_u2} is $\frac{\rho \sqrt{c}}{c - 1}$-Lipschitz w.r.t. the first argument and bounded by $(1 + \frac{1}{c - 1})B$.
            \item the surrogate loss $L_{u_3} (f(\mathbf{x}), \mathbf{y})$ in Eq.\eqref{eq:ral_surrogate_u3} is $2 \rho$-Lipschitz w.r.t. the first argument and bounded by $2B$.
            \item the surrogate loss $L_{u_4} (f(\mathbf{x}), \mathbf{y})$ in Eq.\eqref{eq:ral_surrogate_u4} is $\rho \sqrt{c}$-Lipschitz w.r.t. the first argument and bounded by $cB$.
        \end{enumerate}
    \end{lem}
    Next, we analyze the relationship between true and surrogate losses as follows, which is used for the proof of learning guarantees of algorithms.
    \begin{lem}[The relationship between true and surrogate losses]
        For the ranking loss and its surrogate losses, the following inequalities hold:
        \begin{align}
            & L_r^{0/1} (f(\mathbf{x}), \mathbf{y}) \leq  L_{u_4} (f(\mathbf{x}), \mathbf{y}) \leq c L_{u_2} (f(\mathbf{x}), \mathbf{y}), \\
            & L_r^{0/1} (f(\mathbf{x}), \mathbf{y}) \leq  L_{u_3} (f(\mathbf{x}), \mathbf{y}). 
        \end{align}
    \end{lem}
    The full proof is in Appendix A.2. From this lemma, we can observe that when a learning algorithm minimizes $L_{u_2}$, it also optimizes an upper bound of $L_r^{0/1}$ which depends on $O(c)$. Besides, $L_{u_3}$ and $L_{u_4}$ strictly upper bound $L_r^{0/1}$. These upper bounds of $L_r^{0/1}$ would help to give learning guarantees of corresponding learning algorithms w.r.t. the (partial) ranking loss in the subsequent analyses.
    
    First, we analyze the learning guarantee of $\mathcal{A}^{u_2}$, as follows.
	\begin{thm}[Learning guarantee of $\mathcal{A}^{u_2}$]
		Assume the loss $L_{\phi} =  c L_{u_2}$, where $L_{u_2}$ is defined in Eq.\eqref{eq:ral_surrogate_u2}. Besides, \emph{Assumption~\ref{assump_1}} is satisfied. Then, for any $\delta > 0$, with probability at least $1 - \delta$ over $S$, the following generalization bound holds for all $f \in \mathcal{F}$:
		\begin{align}
			& R_{0/1}^{pr} (f) \leq R_{0/1}^r (f) \leq c \hat{R}^{u_2}_S(f) + 2 \sqrt{2} \rho c (1 +  \frac{1}{c-1}) \sqrt{\frac{\Lambda^2 r^2}{n}} + 3Bc (1+\frac{1}{c-1}) \sqrt{\frac{\log \frac{2}{\delta}}{2n}}.
		\end{align}
	\end{thm}
   The full proof is in Appendix A.3.1. From this theorem, we can see that the learning algorithm $\mathcal{A}^{u_2}$ has a learning guarantee in terms of (partial) ranking loss which depends on $O(c)$.
   
   Then, we provide the learning guarantee of $\mathcal{A}^{u_3}$ in the following theorem.
	\begin{thm}[Learning guarantee of $\mathcal{A}^{u_3}$]
		Assume the loss $L_{\phi} =  L_{u_3}$, where $L_{u_3}$ is defined in Eq.\eqref{eq:ral_surrogate_u3}. Besides, \emph{Assumption~\ref{assump_1}} is satisfied. Then, for any $\delta > 0$, with probability at least $1 - \delta$ over $S$, the following generalization bound holds for all $f \in \mathcal{F}$:
		\begin{align}
			R_{0/1}^{pr} (f) \leq R_{0/1}^r (f) \leq & \hat{R}^{u_3}_S(f) + 4 \sqrt{2} \rho \sqrt{\frac{c \Lambda^2 r^2}{n}} + 6B \sqrt{\frac{\log \frac{2}{\delta}}{2n}}.
		\end{align}
	\end{thm}
    The full proof is in Appendix A.3.2. From this theorem, remarkably, we can see that the learning algorithm $\mathcal{A}^{u_3}$ has a learning guarantee in terms of (partial) ranking loss which depends on $O(\sqrt{c})$, which enjoys the same order as the algorithm $\mathcal{A}^{pa}$ \cite{wu2020multi}.
    
    Finally, we give the learning guarantee of $\mathcal{A}^{u_4}$ as following.
	\begin{thm}[Learning guarantee of $\mathcal{A}^{u_4}$]
		Assume the loss $L_{\phi} =  L_{u_4}$, where $L_{u_4}$ is defined in Eq.\eqref{eq:ral_surrogate_u4}. Besides, \emph{Assumption~\ref{assump_1}} is satisfied. Then, for any $\delta > 0$, with probability at least $1 - \delta$ over $S$, the following generalization bound holds for all $f \in \mathcal{F}$:
		\begin{align}
			R_{0/1}^{pr} (f) \leq R_{0/1}^r (f) \leq & \hat{R}^{u_4}_S(f) + 2 \sqrt{2} \rho c \sqrt{\frac{\Lambda^2 r^2}{n}}  + 3c B \sqrt{\frac{\log \frac{2}{\delta}}{2n}}.
		\end{align}
	\end{thm}
	The full proof is in Appendix A.3.3. The above theorem indicates that $\mathcal{A}^{u_4}$ has a learning guarantee w.r.t. (partial) ranking loss depending on $O(c)$, which is the same as $\mathcal{A}^{u_2}$.

\subsection{Consistency Analyses}

\begin{table}[t]
\scriptsize
\centering
\caption{The penalties of the specific univariate losses in Section~\ref{sec:loss} w.r.t.  the general reweighted form in Eq.~\eqref{eqn:gen_uni}.}
\label{penalty-table}
\begin{center}
\begin{small}
\begin{tabular}{lccc}
\toprule
Loss & $1 / \beta_{\mathbf{y}}^+$ & $1 / \beta_{\mathbf{y}}^-$   \\
\midrule
$L_{u_1}$ & $ {c}$ & $ {c}$    \\
\specialrule{0em}{1pt}{1pt}
$L_{u_2}$ & $ {  |S_{\mathbf{y}}^+| |S_{\mathbf{y}}^-| }$ & $ {  |S_{\mathbf{y}}^+| |S_{\mathbf{y}}^-| }$ \\ 
\specialrule{0em}{1pt}{1pt}
$L_{u_3}$ & $ {|S_{\mathbf{y}}^+|}$  & $ {|S_{\mathbf{y}}^-|}$ \\ 
\specialrule{0em}{1pt}{1pt}
$L_{u_4}$ & $ { \min\{|S_{\mathbf{y}}^+|, |S_{\mathbf{y}}^-|\} }$ & ${ \min\{|S_{\mathbf{y}}^+|, |S_{\mathbf{y}}^-|\}}$    \\
\bottomrule
\end{tabular}
\end{small}
\end{center}
\vskip -0.1in
\end{table}

For consistency, following~\cite{dembczynski2012consistent,gao2013consistency}, we consider the {\it general ranking loss} and the {\it general partial ranking loss} as follows:
\begin{align}
L^{0/1}_{gr}(f(\mathbf{x}), \mathbf{y}) =    \alpha_\mathbf{y}\sum_{(p ,q) \in S_{\mathbf{y}}^+  \times S_{\mathbf{y}}^-} \bigg [ [\![ f_p(\mathbf{x}) \le  f_q(\mathbf{x}) ]\!] \bigg ]  , \label{eqn:cons_rnk_notpartial}
\end{align} and
\begin{align}
L^{0/1}_{gpr}(f(\mathbf{x}), \mathbf{y}) = &    \alpha_\mathbf{y}  \sum_{(p ,q) \in S_{\mathbf{y}}^+  \times S_{\mathbf{y}}^-} \bigg [ [\![ f_p(\mathbf{x}) < f_q(\mathbf{x}) ]\!]
+ \frac{1}{2} [\![ f_p(\mathbf{x} ) = f_q(\mathbf{x} ) ]\!] \bigg ], \label{eqn:cons_rnk}
\end{align} 
where $\alpha_\mathbf{y}$ is a positive penalty. The losses in Eq.~\eqref{eqn:spe_rl} and  Eq.~\eqref{eqn:spe_prl} are special cases of Eq.~\eqref{eqn:cons_rnk_notpartial} and 
Eq.~\eqref{eqn:cons_rnk} with $\alpha_\mathbf{y}  = \frac{1}{ |S_{\mathbf{y}}^+ | |S_{\mathbf{y}}^- | }$ respectively. For clarity and generality, we define the {\it general reweighted univariate surrogate loss} as follows:
\begin{align}
L_{u} (f(\mathbf{x}), \mathbf{y})
= &  		
\sum_{j=1}^c  ([\![ y_j = + 1 ]\!] \beta_\mathbf{y}^+ + [\![ y_j  = - 1 ]\!] \beta_\mathbf{y}^- )  \ell (y_j f_j(\mathbf{x})),
\label{eqn:gen_uni}
\end{align}
where $\beta_\mathbf{y}^+$ and $\beta_\mathbf{y}^-$ are penalties for the positive and negative labels respectively. We assume $\beta_\mathbf{y}^+\beta_\mathbf{y}^- > 0$ for convenience in our analyses. Note that the penalties can be different and all univariate surrogate losses presented in Section~\ref{sec:loss} are special cases of Eq.~\eqref{eqn:gen_uni}. (See Table~\ref{penalty-table} for details.)

Let $\mathcal{B}_L(\mathbf{x}, P(\mathbf{y | \mathbf{x}}))$ denote the set of the Bayes predictors of a loss $L$ given a data point $\mathbf{x}$ and a conditional distribution $P(\mathbf{y | \mathbf{x}})$.  Remarkably, a sufficient and necessary condition (called \emph{multi-label consistency}~\cite{gao2013consistency}) for a surrogate loss to be (Fisher) consistent  w.r.t.  the (partial) ranking loss is presented in the following Lemma~\ref{cons_cond}.
\begin{lem}[Multi-label consistency \cite{gao2013consistency}]\label{cons_cond}
	A surrogate loss $L$  is consistent w.r.t.  a $0/1$ loss $L^{0/1}$, including the general ranking loss in Eq.~\eqref{eqn:cons_rnk_notpartial} and the general partial ranking loss in Eq.~\eqref{eqn:cons_rnk}, if and only if $\forall \mathbf{x}$ and  $P(\mathbf{y} | \mathbf{x})$, $
		\mathcal{B}_{L}(\mathbf{x}, P(\mathbf{y | \mathbf{x}})) \subset 	\mathcal{B}_{L^{0/1}}(\mathbf{x}, P(\mathbf{y | \mathbf{x}}))$.
\end{lem}
 
Note that it takes additional efforts to check the consistency of a new surrogate loss  according to Lemma~\ref{cons_cond},  because one has to enumerate all possible conditional distributions. For the general loss in Eq.~\eqref{eqn:gen_uni}, 
we present more intuitive characterization that only involves the penalties in  Theorem~\ref{thm:cons_log} and Proposition~\ref{cons_pro_hinge},  considering different base losses. For clarity, we refer the readers to Appendix B for all proof.

\begin{thm}[Necessary condition for the consistency of Eq.~\eqref{eqn:gen_uni} w.r.t. Eq.~\eqref{eqn:cons_rnk} with exponential, logistic or squared hinge loss]
A general reweighted univariate surrogate loss  in Eq.~\eqref{eqn:gen_uni} with $\ell(z) = e^{-z}$, $\ell(z) = \ln (1 + e^{-z})$ or $\ell(z) = (\max\{0, 1- z\})^2$ is consistent w.r.t.  the general partial ranking loss in Eq.~\eqref{eqn:cons_rnk}  only if $\exists \tau  > 0$, $\beta_\mathbf{y}^+ \beta_\mathbf{y}^-  = \tau \alpha_\mathbf{y}^2  $ for all $\mathbf{y}$ such that  $1-c\le \sum_{1\le j \le c}y_j \le c-1$. 
\label{thm:cons_log}
\end{thm}

Note that, when $c \le 3$, the penalties of $L_{u_1}$, $L_{u_3}$ and $L_{u_4}$ may coincide with that of $L_{u_2}$ up to a multiplicative constant.
When $c \ge 4$, it is straightforward to construct counter examples that violate the necessary condition in Theorem~\ref{thm:cons_log} and obtain the following Corollary~\ref{cons_cor_log_all}.
\begin{corollary}[Inconsistency of $L_{u_1}$, $L_{u_3}$ and $L_{u_4}$ w.r.t. Eq.~\eqref{eqn:spe_prl} with exponential, logistic or squared hinge loss]
If $c \ge 4$, $L_{u_1}$, $L_{u_3}$ and $L_{u_4}$ with  $\ell(z) = e^{-z}$ or $\ell(z) = \ln (1 + e^{-z})$ or $\ell(z) = (\max\{0, 1- z\})^2$
 are inconsistent w.r.t.  the partial ranking loss in Eq.~\eqref{eqn:spe_prl}. \label{cons_cor_log_all}
\end{corollary}

We further show the inconsistency of the general reweighted univariate loss  in Eq.~\eqref{eqn:gen_uni} w.r.t.  the general partial ranking loss in Eq.~\eqref{eqn:cons_rnk}  with hinge loss. Note that this includes the inconsistency of $L_{u_1}$, $L_{u_3}$ and $L_{u_4}$ w.r.t. Eq.~\eqref{eqn:spe_prl}.

\begin{pro}[Inconsistency of Eq.~\eqref{eqn:gen_uni} w.r.t. Eq.~\eqref{eqn:cons_rnk} with hinge loss]
The general reweighted univariate surrogate loss  in Eq.~\eqref{eqn:gen_uni} with $\ell(z) = \max \{0, 1 - z\}$  are inconsistent w.r.t.  the general partial ranking loss in Eq.~\eqref{eqn:cons_rnk}, for all positive penalties $\alpha_{\mathbf{y}}$, $\beta^+_{\mathbf{y}}$ and $\beta^-_{\mathbf{y}}$.  \label{cons_pro_hinge}
\end{pro}

An immediate conclusion from Corollary~\ref{cons_cor_log_all} and Proposition~\ref{cons_pro_hinge} is that $L_{u_1}$, $L_{u_3}$ and $L_{u_4}$ are inconsistent w.r.t.  the ranking loss in Eq.~\eqref{eqn:spe_rl} because $\mathcal{B}_{L_{r}^{0/1}}(\mathbf{x}, P(\mathbf{y | \mathbf{x}})) \subset 	\mathcal{B}_{L_{pr}^{0/1}}(\mathbf{x}, P(\mathbf{y | \mathbf{x}}))$~\cite{gao2013consistency}.
Compared to existing work~\cite{dembczynski2012consistent,gao2013consistency}, although  Theorem~\ref{thm:cons_log} and Proposition~\ref{cons_pro_hinge} are negative, such results consider surrogate losses in a more general reweighted form, i.e. Eq.~\eqref{eqn:gen_uni}, which may be of independent interest.

\section{Experiments}
\label{sec:experimental_results}

\begin{table}[t]
\scriptsize
\caption{Basic statistics of the benchmark datasets.}
\label{tab:datasets}
\vskip 0.15in
\begin{center}
\begin{small}
\begin{tabular}{lcccc}
\toprule
Dataset & \#Instance & \#Feature & \#Label & Domain \\
\midrule
emotions & 593 & 72 & 6 & music \\
image & 2000 & 294 & 5 & images \\
scene & 2407 & 294 & 6 & images \\
yeast & 2417 & 103 & 14 & biology \\
enron & 1702 & 1001 & 53 & text \\
rcv1-subset1 & 6000 & 944 & 101 & text \\
bibtex & 7395 & 1836 & 159 & text \\
corel5k & 5000 & 499 & 374 & images \\
mediamill & 43907 & 120 & 101 & video \\
delicious & 16105 & 500 & 983 & text(web) \\
\bottomrule
\end{tabular}
\end{small}
\end{center}
\vskip -0.1in
\end{table}

To validate our theory findings, 
we evaluate all algorithms presented in Section~\ref{sec:methods}  on $10$ widely-used benchmark datasets\footnote{Datasets are available at http://mulan.sourceforge.net/datasets-mlc.html and \\ http://palm.seu.edu.cn/zhangml/.} 
with various domains and sizes of label and data. We summarize their statistics in  Table~\ref{tab:datasets}. Since the first four datasets are not properly prepossessed, we normalize the input as zero mean and unit variance following~\cite{wu2020multi}. 
For all the learning algorithms, we utilize the linear models with the base logistic loss for simplicity and a fair comparison. Besides, we use the same efficient stochastic algorithm (i.e. SVRG-BB~\cite{tan2016barzilai}) to solve these convex optimization problems.  Moreover, for fairness, we take $3$-fold cross validation on each dataset, where the hyper-parameter $\lambda$ is searched in a wide range of $\{10^{-8}, 10^{-7}, \cdots, 10^2\}$ for all algorithms. We use the ranking loss as the evaluation measure.

The experimental results are summarized in Table~\ref{tab:benchmark_results} and we refer the readers to  Appendix C for complete results with standard deviations. First, we observe that $\mathcal{A}^{pa}$ and $\mathcal{A}^{u_3}$ outperform the others especially $\mathcal{A}^{u_2}$  on almost all benchmarks. It agrees with our generalization analyses: $\mathcal{A}^{pa}$ and $\mathcal{A}^{u_3}$ enjoy a generalization error bound of $O(\sqrt{c})$ while the others have a bound of $O(c)$. Besides, we would like to emphasize that such results do not contradict with the consistency results. In fact, there are two assumptions in consistency analyses are violated in the real settings. The first one is that the Bayes predictor may not be linear and the second one is that the number of samples may not be sufficient to achieve the Bayes predictor, which explain the relatively weaker results of $\mathcal{A}^{u_2}$ considering its consistency. In this sense, the generalization error bounds may provide more insights than consistency when the number of training samples is finite (or not sufficiently large) and the hypothesis space is not realizable.

Further, we also note that $\mathcal{A}^{u_3}$ outperforms $\mathcal{A}^{pa}$ on the last four datasets with a relatively large $c$. The underlying mechanism is not clear yet. Our hypothesis is that our univariate loss is easier to optimize than the pairwise loss, which may provide additional benefits beyond the scope of the generalization analyses. A deeper analysis is left as future work.

Moreover, as for computation efficiency, the pairwise loss is much slower than all univariate ones, including $\mathcal{A}^{u_3}$. 
Indeed, $\mathcal{A}^{pa}$ takes more than a week using a 48-core CPU server on the delicious dataset with $c=983$ and we do not finish it. We provide quantitative results of the running time in Appendix C.

In conclusion, the benchmark results 
show the promise of our $L_{u_3}$ in terms of both efficiency and effectiveness.

\begin{table}[t]
\renewcommand\tabcolsep{4pt}
\scriptsize
\caption{Ranking loss of all five algorithms on benchmark datasets. On each dataset, the top two algorithms are highlighted in bold and the top one is labeled with $^{\dagger}$. }
\label{tab:benchmark_results}
\vskip 0.15in
\begin{center}
\begin{small}
\begin{tabular}{lccccc}
\toprule
Dataset & $\mathcal{A}^{pa}$ & $\mathcal{A}^{u_1}$ & $\mathcal{A}^{u_2}$ & $\mathcal{A}^{u_3}$ & $\mathcal{A}^{u_4}$ \\
\midrule
emotions & $\bf 0.1511^{\dagger}$ & $0.1538$ & $0.1587$ & $\bf0.1530$ & $0.1616$ \\
image & $\bf 0.1625^{\dagger}$ & $0.1642$ & $0.1653$ & $\bf0.1645$ & $0.1678$ \\
scene & $\bf 0.0696^{\dagger}$ & $0.0809$ & $0.0821$ & $\bf0.0768$ & $0.0806$ \\
yeast & $\bf 0.1766^{\dagger}$ & $0.1768$ & $0.1785$ & $\bf0.1767$ & $0.1816$ \\
enron & $\bf 0.0682^{\dagger}$ & $0.0724$ & $\bf0.0696$ & $0.0698$ & $0.0715$ \\
rcv1-subset1 & $\bf 0.0361^{\dagger}$ & $0.0418$ & $0.0392$ & $\bf0.0368$ & $0.0391$ \\
bibtex & $\bf0.0516$ & $0.0545$ & $0.0551$ & $\bf 0.0401^{\dagger}$ & $0.0538$ \\
corel5k & $\bf0.1081$ & $0.1091$ & $0.1099$ & $\bf 0.1063^{\dagger}$ & $0.1096$ \\
mediamill & $\bf0.0395$ & $0.0402$ & $0.0412$ & $\bf 0.0389^{\dagger}$ & $0.0405$ \\
delicious & - & $\bf0.0960$ & $0.0974$ & $\bf 0.0946^{\dagger}$ & $0.0978$ \\
\bottomrule
\end{tabular}
\end{small}
\end{center}
\vskip -0.1in
\end{table}

\section{Conclusion and Discussion}
\label{sec:conclusion_future_work}

This paper presents a systematic study from two complementary perspectives of consistency and generalization error bounds in multi-label ranking. In particular, existing consistent univariate losses lead to an error bound depending on $O(c)$ while the inconsistent pairwise losses enjoy an error bound of $O(\sqrt{c})$~\cite{wu2020multi}. 
Inspired by the generalization analyses, we present two reweighted surrogate univariate losses that strictly upper bound the (partial) ranking loss.
Surprisingly, though not consistent, one of them  enjoys an error bound depending on $O(\sqrt{c})$, which is nearly the same as the pairwise loss and retains the computational efficiency. Empirical results validate our theory findings.

Some problems in MLR and MLC are still open and may inspire future work. Theoretically, consistency provides valuable insights when learning on  large-scale samples with a unconstrained function space. However, our empirical results show that generalization explains the behaviour of a loss more accurately than consistency if the two assumptions are violated. Generally, a deeper understanding of the complementary roles of the two perspectives is intriguing.
Besides, it is also attractive to investigate whether one can design a loss such that it is consistent and its corresponding learning algorithm has a tight generalization bound.

\bibliographystyle{unsrt} 
\bibliography{references} 

\clearpage

\appendix

\section{Generalization analyses}
\subsection{Proof of Lemma 1}
    \begin{lem_appendix}{1}[The properties of surrogate losses]
        Assume that the base (convex) loss $\ell (z)$ is $\rho$-Lipschitz continuous and bounded by $B$. Then, the following holds. 
        
        (1) the surrogate loss $L_{u_2} (f(\mathbf{x}), \mathbf{y})$ in Eq.(11) is $\frac{\rho \sqrt{c}}{c - 1}$-Lipschitz w.r.t. the first argument and bounded by $(1 + \frac{1}{c - 1})B$.
        
        (2) the surrogate loss $L_{u_3} (f(\mathbf{x}), \mathbf{y})$ in Eq.(12) is $2 \rho$-Lipschitz w.r.t. the first argument and bounded by $2B$.
        
        (3) the surrogate loss $L_{u_4} (f(\mathbf{x}), \mathbf{y})$ in Eq.(13) is $\rho \sqrt{c}$-Lipschitz w.r.t. the first argument and bounded by $cB$.
    \end{lem_appendix}
    \begin{proof}
        (1) For the surrogate univariate loss $L_{u_2}(f(\mathbf{x}), \mathbf{y})$, $\forall f^1, f^2 \in \mathcal{F}$, the following holds:
            \begin{align*}
    			& \quad | L_{u_2}(f^1, \mathbf{y}) - L_{u_2}(f^2, \mathbf{y}) | \\
    			& = \frac{1}{|S_{\mathbf{y}}^+| |S_{\mathbf{y}}^-|} \sum_{j=1}^c | \ell (y_j f^1_j) - \ell (y_j f^2_j) | \\
    			& \leq \frac{1}{|S_{\mathbf{y}}^+| |S_{\mathbf{y}}^-|} \sum_{j=1}^c \rho | y_j f^1_j - y_j f^2_j | \qquad \qquad \quad (\ell(z) \ is \ \rho-Lipschitz) \\
    			& \leq \frac{\rho c}{|S_{\mathbf{y}}^+| |S_{\mathbf{y}}^-|} \bigg [ \frac{1}{c} \sum_{j=1}^c | f^1_j - f^2_j |^2 \bigg ]^{1/2} \qquad \qquad \quad (Jensen's \ Inequality) \\
    			& = \frac{\rho \sqrt{c}}{|S_{\mathbf{y}}^+| |S_{\mathbf{y}}^-|} \| f^1 - f^2 \| \\
    			& \leq \frac{\rho \sqrt{c}}{c - 1} \| f^1 - f^2 \| \qquad \qquad \quad (c - 1 \leq |S_{\mathbf{y}}^+| |S_{\mathbf{y}}^-| \leq \frac{c^2}{4}).
    	    \end{align*}
    	    Since the inequality $c - 1 \leq |S_{\mathbf{y}}^+| |S_{\mathbf{y}}^-| \leq \frac{c^2}{4}$ holds, it is easy to check that $L_{u_2}$ is bounded by $(1 + \frac{1}{c - 1})B$.
    	    
        (2) For the surrogate univariate loss $L_{u_3}(f(\mathbf{x}), \mathbf{y})$, $\forall f^1, f^2 \in \mathcal{F}$, the following holds:
    	    \begin{align*}
    			& \quad | L_{u_3}(f^1, \mathbf{y}) - L_{u_3}(f^2, \mathbf{y}) | \\
    			& = \Big | \frac{\sum_{p \in S_{\mathbf{y}}^+} [\ell(y_p^1 f_p^1) - \ell(y_p^2 f_p^2)]}{|S_{\mathbf{y}}^+|} + \frac{\sum_{q \in S_{\mathbf{y}}^-} [\ell(y_q^1 f_q^1) - \ell(y_q^2 f_q^2)]}{|S_{\mathbf{y}}^-|} \Big | \\
    			& \leq \frac{\sum_{p \in S_{\mathbf{y}}^+} | \ell(f_p^1) - \ell(f_p^2) |}{|S_{\mathbf{y}}^+|} +  \frac{\sum_{q \in S_{\mathbf{y}}^-} |\ell(-f_q^1) - \ell(-f_q^2)|}{|S_{\mathbf{y}}^-|}     \qquad \quad (|a + b| \leq |a| + |b|) \\
    			& \leq \frac{\sum_{p \in S_{\mathbf{y}}^+} \rho | f_p^1 - f_p^2 |}{|S_{\mathbf{y}}^+|} +  \frac{\sum_{q \in S_{\mathbf{y}}^-} \rho | f_q^1 - f_q^2 |}{|S_{\mathbf{y}}^-|} \qquad \qquad \qquad \quad (\ell(z) \ is \ \rho-Lipschitz) \\ 
    			& \leq \rho \bigg [ \frac{\sum_{p \in S_{\mathbf{y}}^+} | f_p^1 - f_p^2 |^2}{|S_{\mathbf{y}}^+|} \bigg ]^{1/2} + \rho \bigg [ \frac{\sum_{q \in S_{\mathbf{y}}^-} | f_q^1 - f_q^2 |^2}{|S_{\mathbf{y}}^-|} \bigg ]^{1/2} \qquad (Jensen's \ Inequality) \\
    			& \leq \rho \bigg [ \frac{\sum_{p \in S_{\mathbf{y}}^+} | f_p^1 - f_p^2 |^2 + \sum_{q \in S_{\mathbf{y}}^-} | f_q^1 - f_q^2 |^2}{|S_{\mathbf{y}}^+|} \bigg ]^{1/2} + \rho \bigg [ \frac{\sum_{p \in S_{\mathbf{y}}^+} | f_p^1 - f_p^2 |^2 + \sum_{q \in S_{\mathbf{y}}^-} | f_q^1 - f_q^2 |^2}{|S_{\mathbf{y}}^-|} \bigg ]^{1/2} \\
    			& \leq 2 \rho \bigg [ \frac{\sum_{j=1}^c | f_j^1 - f_j^2 |^2}{\min\{ |S_{\mathbf{y}}^+|, |S_{\mathbf{y}}^-| \}} \bigg ]^{1/2} \\
    			& \leq 2 \rho \| f^1 - f^2 \| \qquad \qquad \qquad \qquad \qquad \quad (1 \leq \min\{|S_{\mathbf{y}}^+|, |S_{\mathbf{y}}^-|\} \leq \frac{c}{2}).
    	    \end{align*}
    	    It is easy to check that $L_{u_3}$ is bounded by $2B$.
    	    
        (3) For the surrogate univariate loss $L_{u_4}(f(\mathbf{x}), \mathbf{y})$, $\forall f^1, f^2 \in \mathcal{F}$, the following holds:
            \begin{align*}
    			& \quad | L_{u_4}(f^1, \mathbf{y}) - L_{u_4}(f^2, \mathbf{y}) | \\
    			& = \frac{1}{\min\{|S_{\mathbf{y}}^+|, |S_{\mathbf{y}}^-|\}} \sum_{j=1}^c | \ell (y_j f^1_j) - \ell (y_j f^2_j) | \\
    			& \leq \frac{1}{\min\{|S_{\mathbf{y}}^+|, |S_{\mathbf{y}}^-|\}} \sum_{j=1}^c \rho | y_j f^1_j - y_j f^2_j | \qquad \qquad \quad (\ell(z) \ is \ \rho-Lipschitz) \\
    			& \leq \frac{\rho c}{\min\{|S_{\mathbf{y}}^+|, |S_{\mathbf{y}}^-|\}} \bigg [ \frac{1}{c} \sum_{j=1}^c | f^1_j - f^2_j |^2 \bigg ]^{1/2} \qquad \qquad \quad (Jensen's \ Inequality) \\
    			& = \frac{\rho \sqrt{c}}{\min\{|S_{\mathbf{y}}^+|, |S_{\mathbf{y}}^-|\}} \| f^1 - f^2 \| \\
    			& \leq \rho \sqrt{c} \| f^1 - f^2 \| \qquad \qquad \quad (1 \leq \min\{|S_{\mathbf{y}}^+|, |S_{\mathbf{y}}^-|\} \leq \frac{c}{2}).
    	    \end{align*}
    	    Since the inequality $1 \leq \min\{|S_{\mathbf{y}}^+|, |S_{\mathbf{y}}^-|\} \leq \frac{c}{2}$ holds, it is easy to check that $L_{u_4}$ is bounded by $cB$.
    \end{proof}

\subsection{Proof of Lemma 2}
    \begin{lem_appendix}{2}[The relationship between true and surrogate losses]
        For the ranking loss and its surrogate losses, the following inequalities hold:
        \begin{align}
            & L_r^{0/1} (f(\mathbf{x}), \mathbf{y}) \leq  L_{u_4} (f(\mathbf{x}), \mathbf{y}) \leq c L_{u_2} (f(\mathbf{x}), \mathbf{y}), \\
            & L_r^{0/1} (f(\mathbf{x}), \mathbf{y}) \leq  L_{u_3} (f(\mathbf{x}), \mathbf{y}). 
        \end{align}
    \end{lem_appendix}
    \begin{proof}
        For the first inequality, the following holds:
    	\begin{align*}
    			& L_r^{0/1} (f(\mathbf{x}), \mathbf{y}) \leq L_r^{0/1} (sgn \circ f(\mathbf{x}), \mathbf{y}) \\
    			& \qquad \qquad \qquad = \frac{1}{|S_{\mathbf{y}}^+||S_{\mathbf{y}}^-|} \sum_{p \in S_{\mathbf{y}}^+} \sum_{q \in S_{\mathbf{y}}^-}  [\![ sgn(f_p(\mathbf{x})) \leq sgn(f_q(\mathbf{x})) ]\!] \\
    			& \qquad \qquad \qquad = \frac{1}{|S_{\mathbf{y}}^+| |S_{\mathbf{y}}^-|} \bigg [ |S_{\mathbf{y}}^-| \sum_{p \in S_{\mathbf{y}}^+} [\![ sgn (f_p(\mathbf{x}))  \neq 1 ]\!] + |S_{\mathbf{y}}^+| \sum_{q \in S_{\mathbf{y}}^-} [\![ sgn (f_q(\mathbf{x}))  \neq -1 ]\!] - \\
    			& \qquad \qquad \qquad \qquad \qquad \bigg \{ \sum_{p \in S_{\mathbf{y}}^+} [\![ sgn (f_p(\mathbf{x}))  \neq 1 ]\!] \bigg \} \bigg \{ \sum_{q \in S_{\mathbf{y}}^-} [\![ sgn (f_q(\mathbf{x}))  \neq -1 ]\!] \bigg \} \bigg ] \\
    			& \qquad \qquad \qquad \leq \frac{1}{|S_{\mathbf{y}}^+| |S_{\mathbf{y}}^-|} \bigg [ |S_{\mathbf{y}}^-| \sum_{p \in S_{\mathbf{y}}^+} [\![ sgn (f_p(\mathbf{x}))  \neq 1 ]\!] + |S_{\mathbf{y}}^+| \sum_{q \in S_{\mathbf{y}}^-} [\![ sgn (f_q(\mathbf{x}))  \neq -1 ]\!] \bigg ] \\
    			& \qquad \qquad \qquad \leq \frac{\max \{|S_{\mathbf{y}}^+|, |S_{\mathbf{y}}^-|\}}{|S_{\mathbf{y}}^+| |S_{\mathbf{y}}^-|} \bigg [ \sum_{p \in S_{\mathbf{y}}^+} [\![ sgn (f_p(\mathbf{x}))  \neq 1 ]\!] + \sum_{q \in S_{\mathbf{y}}^-} [\![ sgn (f_q(\mathbf{x}))  \neq -1 ]\!] \bigg ] \\
    			& \qquad \qquad \qquad = \frac{\max \{|S_{\mathbf{y}}^+|, |S_{\mathbf{y}}^-|\}}{|S_{\mathbf{y}}^+| |S_{\mathbf{y}}^-|} \sum_{j=1}^c [\![ sgn (f_j(\mathbf{x}))  \neq y_j ]\!] \\
    			& \qquad \qquad \qquad \leq \frac{\max \{|S_{\mathbf{y}}^+|, |S_{\mathbf{y}}^-|\}}{|S_{\mathbf{y}}^+| |S_{\mathbf{y}}^-|} \sum_{j=1}^c \ell (y_j f_j(\mathbf{x}))   \\
    			& \qquad \qquad \qquad = \frac{1}{\min\{|S_{\mathbf{y}}^+|,|S_{\mathbf{y}}^-|\}} \sum_{j=1}^c \ell (y_j f_j(\mathbf{x})) \\
    			& \qquad \qquad \qquad =  L_{u_4} (f(\mathbf{x}), \mathbf{y})  \\
    			& \qquad \qquad \qquad \leq c L_{u_2} (f(\mathbf{x}), \mathbf{y}) \qquad \qquad \qquad (\frac{c}{2} \leq \max \{|S_{\mathbf{y}}^+|, |S_{\mathbf{y}}^-|\} \leq c - 1).
    	\end{align*}
    
        For the second inequality, the following holds:
        \begin{align*}
        		& L_r^{0/1} (f(\mathbf{x}), \mathbf{y}) \leq L_r^{0/1} (sgn \circ f(\mathbf{x}), \mathbf{y}) \\
        		& \qquad \qquad \qquad = \frac{1}{|S_{\mathbf{y}}^+||S_{\mathbf{y}}^-|} \sum_{p \in S_{\mathbf{y}}^+} \sum_{q \in S_{\mathbf{y}}^-}  [\![ sgn(f_p(\mathbf{x})) \leq sgn(f_q(\mathbf{x})) ]\!] \\
        		& \qquad \qquad \qquad = \frac{1}{|S_{\mathbf{y}}^+| |S_{\mathbf{y}}^-|} \bigg [ |S_{\mathbf{y}}^-| \sum_{p \in S_{\mathbf{y}}^+} [\![ sgn (f_p(\mathbf{x}_i))  \neq 1 ]\!] + |S_{\mathbf{y}}^+| \sum_{q \in S_{\mathbf{y}}^-} [\![ sgn (f_q(\mathbf{x}_i))  \neq -1 ]\!] - \\
        		& \qquad \qquad \qquad \qquad \qquad \bigg \{ \sum_{p \in S_{\mathbf{y}}^+} [\![ sgn (f_p(\mathbf{x}))  \neq 1 ]\!] \bigg \} \bigg \{ \sum_{q \in S_{\mathbf{y}}^-} [\![ sgn (f_q(\mathbf{x}))  \neq -1 ]\!] \bigg \} \bigg ] \\
        		& \qquad \qquad \qquad \leq \frac{1}{|S_{\mathbf{y}}^+| |S_{\mathbf{y}}^-|} \bigg [ |S_{\mathbf{y}}^-| \sum_{p \in S_{\mathbf{y}}^+} [\![ sgn (f_p(\mathbf{x}))  \neq 1 ]\!] + |S_{\mathbf{y}}^+| \sum_{q \in S_{\mathbf{y}}^-} [\![ sgn (f_q(\mathbf{x}))  \neq -1 ]\!] \bigg ] \\
        		& \qquad \qquad \qquad = \frac{\sum_{p \in S_{\mathbf{y}}^+} [\![ sgn (f_p(\mathbf{x}))  \neq 1 ]\!] }{|S_{\mathbf{y}}^+|} + \frac{\sum_{q \in S_{\mathbf{y}}^-} [\![ sgn (f_q(\mathbf{x}))  \neq -1 ]\!] }{|S_{\mathbf{y}}^-|} \\
        		& \qquad \qquad \qquad \leq L_{u_3} (f(\mathbf{x}), \mathbf{y}).
        \end{align*}
	\end{proof}

\subsection{Proof of Theorem 1, 2 and 3}
    Following \cite{wu2020multi}, we also give the base theorem used in the subsequent generalization analysis, as follows.
\begin{thm_appendix} {A.1} [The base theorem for generalization analysis~\cite{wu2020multi}]
\label{thm:base}
	Assume the loss function $L_{\phi}: \mathbb{R}^c \times \{-1, +1\}^c \rightarrow \mathbb{R}_+$ is $\mu$-Lipschitz continuous w.r.t. the first argument and bounded by $M$. Besides, (1) and (2) in \emph{Assumption 1} are satisfied. Then, for any $\delta > 0$, with probability at least $1 - \delta$ over the draw of an i.i.d. sample $S$ of size $n$, the following generalization bound holds for all $f \in \mathcal{F}$:
	\begin{equation}
		R_{\phi}(f) \leq \hat{R}_S(f) + 2 \sqrt{2} \mu \sqrt{\frac{c \Lambda^2 r^2}{n}} + 3M \sqrt{\frac{\log \frac{2}{\delta}}{2n}}.
	\end{equation}
\end{thm_appendix}

    \subsubsection{Proof of Theorem 1}
    	\begin{thm_appendix}{1}[Learning guarantee of $\mathcal{A}^{u_2}$]
    		Assume the loss function $L_{\phi} =  c L_{u_2}$, where $L_{u_2}$ is defined in Eq.(11). Besides, \emph{Assumption 1} is satisfied. Then, for any $\delta > 0$, with probability at least $1 - \delta$ over $S$, the following generalization bound holds for all $f \in \mathcal{F}$:
    		\begin{align}
    			 R_{0/1}^{pr} (f) \leq R_{0/1}^r (f) \leq c \hat{R}^{u_2}_S(f) + 2 \sqrt{2} \rho c (1 +  \frac{1}{c-1}) \sqrt{\frac{\Lambda^2 r^2}{n}} + 3Bc (1+\frac{1}{c-1}) \sqrt{\frac{\log \frac{2}{\delta}}{2n}}.
    		\end{align}
    	\end{thm_appendix}
        \begin{proof}
        	Since $L_{\phi} = c L_{u_2}$, we can get its Lipschitz constant (i.e. $\rho \sqrt{c} (1 + \frac{1}{c - 1})$) and bounded value (i.e. $c(1 + \frac{1}{c - 1}) B$) from (1) in Lemma 1. Then, applying Theorem \ref{thm:base}, we can get that,  for any $\delta > 0$, with probability at least $1 - \delta$ over $S$, the following generalization bound  holds for all $f \in \mathcal{F}$: 
        		\begin{align}
        			R_{\phi}(f) = c R^{u_2}(f) \leq c \hat{R}^{u_2}_S(f) + 2 \sqrt{2} \rho c (1 +  \frac{1}{c-1}) \sqrt{\frac{\Lambda^2 r^2}{n}} + 3Bc (1+\frac{1}{c-1}) \sqrt{\frac{\log \frac{2}{\delta}}{2n}}.
        		\end{align}
        	Besides, from Lemma 2 (i.e. InEq.(20)), we can get the inequality $R_{0/1}^{pr}(f) \leq R_{0/1}^r(f) \leq c R^{u_2}(f)$. Thus, we can get this theorem.
        \end{proof}
    
    \subsubsection{Proof of Theorem 2}
        \begin{thm_appendix}{2}[Learning guarantee of $\mathcal{A}^{u_3}$]
    		Assume the loss function $L_{\phi} =  L_{u_3}$, where $L_{u_3}$ is defined in Eq.(12). Besides, \emph{Assumption 1} is satisfied. Then, for any $\delta > 0$, with probability at least $1 - \delta$ over $S$, the following generalization bound holds for all $f \in \mathcal{F}$:
    		\begin{align}
    			R_{0/1}^{pr} (f) \leq R_{0/1}^r (f) \leq \hat{R}^{u_3}_S(f) + 4 \sqrt{2} \rho \sqrt{\frac{c \Lambda^2 r^2}{n}} + 6B \sqrt{\frac{\log \frac{2}{\delta}}{2n}}.
    		\end{align}
    	\end{thm_appendix}
        \begin{proof}
        	Since $L_{\phi} = L_{u_3}$, we can get its Lipschitz constant (i.e. $2 \rho$) and bounded value (i.e. $2 B$) from (2) in Lemma 1. Then, applying Theorem \ref{thm:base} and the inequality $R_{0/1}^{pr}(f) \leq R_{0/1}^r(f) \leq R^{u_3}(f)$ from Lemma 2 (i.e. InEq.(21)), we can get this theorem.
        \end{proof}
    	
    \subsubsection{Proof of Theorem 3}
    	\begin{thm_appendix}{3}[Learning guarantee of $\mathcal{A}^{u_4}$]
    		Assume the loss function $L_{\phi} =  L_{u_4}$, where $L_{u_4}$ is defined in Eq.(13). Besides, \emph{Assumption 1} is satisfied. Then, for any $\delta > 0$, with probability at least $1 - \delta$ over $S$, the following generalization bound holds for all $f \in \mathcal{F}$:
    		\begin{align}
    			R_{0/1}^{pr} (f) \leq R_{0/1}^r (f) \leq  \hat{R}^{u_4}_S(f) + 2 \sqrt{2} \rho c \sqrt{\frac{\Lambda^2 r^2}{n}} + 3c B \sqrt{\frac{\log \frac{2}{\delta}}{2n}}.
    		\end{align}
    	\end{thm_appendix}
        \begin{proof}
        	Since $L_{\phi} = c L_{u_4}$, we can get its Lipschitz constant (i.e. $\rho \sqrt{c}$) and bounded value (i.e. $c B$) from (1) in Lemma 1. Then, applying Theorem \ref{thm:base} and the inequality $R_{0/1}^{pr}(f) \leq R_{0/1}^r(f) \leq R^{u_4}(f)$ from Lemma 2 (i.e. InEq.(20)), we can get this theorem.
        \end{proof}

\section{Consistency Analyses}

Recall that the ranking loss and the partial ranking loss are defined as 
\begin{align}
    L_r^{0/1}(f(\mathbf{x}), \mathbf{y}) =  \frac{\sum_{(p, q) \in S_{\mathbf{y}}^+ \times S_{\mathbf{y}}^-}  [\![ f_p(\mathbf{x}) \leq f_q(\mathbf{x}) ]\!]}{|S_{\mathbf{y}}^+| |S_{\mathbf{y}}^-|}, \label{eqn:spe_rl_app}
\end{align}
and
\begin{equation}
\begin{split}
    & L_{pr}^{0/1}(f(\mathbf{x}), \mathbf{y}) = \frac{1}{|S_{\mathbf{y}}^+| |S_{\mathbf{y}}^-|} \sum_{(p, q) \in S_{\mathbf{y}}^+ \times S_{\mathbf{y}}^-} \bigg[ [\![ f_p(\mathbf{x}) < f_q(\mathbf{x}) ]\!]
     + \frac{1}{2} [\![ f_p(\mathbf{x}) = f_q(\mathbf{x}) ]\!] \bigg  ], \label{eqn:spe_prl_app}
\end{split}
\end{equation}
respectively. For generality, following~\cite{dembczynski2012consistent,gao2013consistency}, we do not specify the penalties in the losses at beginning. 
Recall that the {\it general ranking loss} is defined as 
\begin{align}
L^{0/1}_{gr}(f(\mathbf{x}), \mathbf{y}) =    \alpha_\mathbf{y}  \sum_{(p ,q) \in S_{\mathbf{y}}^+  \times S_{\mathbf{y}}^-} \bigg [ [\![ f_p(\mathbf{x}) \le  f_q(\mathbf{x}) ]\!] \bigg ], \label{eqn:cons_rnk_notpartial_app}
\end{align}
where $\alpha_\mathbf{y}$ is a positive penalty, and the {\it general partial ranking loss} is in a similar form of 
\begin{align}
L^{0/1}_{gpr}(f(\mathbf{x}), \mathbf{y}) =    \alpha_\mathbf{y}  \sum_{(p ,q) \in S_{\mathbf{y}}^+  \times S_{\mathbf{y}}^-} \bigg [ [\![ f_p(\mathbf{x}) < f_q(\mathbf{x}) ]\!] + \frac{1}{2} [\![ f_p(\mathbf{x} ) = f_q(\mathbf{x} ) ]\!] \bigg ]. \label{eqn:cons_rnk_app}
\end{align}
The commonly used ranking loss and partial ranking loss are the spacial cases of Eq.~\eqref{eqn:cons_rnk_notpartial_app} and 
Eq.~\eqref{eqn:cons_rnk} with $\alpha_\mathbf{y}  = \frac{1}{ |S_{\mathbf{y}}^+ | |S_{\mathbf{y}}^- | }$ respectively. 
Also, recall that the {\it general reweighted univariate surrogate loss} is defined as follows:
\begin{align}
L_{u} (f(\mathbf{x}), \mathbf{y})
& =  		
\sum_{j=1}^c  ([\![ y_j = + 1 ]\!] \beta_\mathbf{y}^+ + [\![ y_j  = - 1 ]\!] \beta_\mathbf{y}^- )  \ell (y_j f_j(\mathbf{x})),
\label{eqn:our_s_loss}
\end{align}
where $\beta_\mathbf{y}^+$ and $\beta_\mathbf{y}^-$ are positive penalties. All univariate surrogate losses mentioned in the main text are spacial cases of Eq.~\eqref{eqn:our_s_loss}, respectively.

Let $\mathcal{B}_L(\mathbf{x}, P(\mathbf{y | \mathbf{x}}))$ denote the set of the Bayes predictors of a loss $L$ given a data point $\mathbf{x}$ and a conditional distribution $P(\mathbf{y | \mathbf{x}})$.  Remarkably, a sufficient and necessary condition (called \emph{multi-label consistency}~\cite{gao2013consistency}) for a surrogate loss to be (Fisher) consistent  w.r.t. the (partial) ranking loss is presented in the following Lemma~\ref{cons_cond_app}. 
\begin{lem_appendix} {3} [Multi-label consistency \cite{gao2013consistency}]\label{cons_cond_app}
	A surrogate loss $L$  is consistent w.r.t.  a $0/1$ loss $L^{0/1}$, including the general ranking loss in Eq.~\eqref{eqn:cons_rnk_notpartial_app} and the general partial ranking loss in Eq.~\eqref{eqn:cons_rnk_app}, if and only if $\forall \mathbf{x}$ and  $P(\mathbf{y} | \mathbf{x})$, $
		\mathcal{B}_{L}(\mathbf{x}, P(\mathbf{y | \mathbf{x}})) \subset 	\mathcal{B}_{L^{0/1}}(\mathbf{x}, P(\mathbf{y | \mathbf{x}}))$.
\end{lem_appendix}

For convenience, we define 
\begin{align}
    \Delta_{pq}^{rk} =\sum_{\mathbf{y}: y_p = s_r,  y_q = s_k}  \alpha_\mathbf{y}P(\mathbf{y}| \mathbf{x}) ,  \textrm{ and }
    \Delta_p^{r} = \sum_{\mathbf{y}: y_p = s_r} \alpha_\mathbf{y} P(\mathbf{y}| \mathbf{x}) = \Delta_{pq}^{r+} +  \Delta_{pq}^{r-}, \forall p \neq q,\label{eqn:cons_pni}
\end{align}
where $r, k \in \{+, -\}$ and $s_+ = +1$ and $s_- = -1$.
The following Lemma~\ref{cons_r_bayes} characterizes the set of the Bayes predictors w.r.t. the general ranking loss in Eq.~\eqref{eqn:cons_rnk_notpartial_app} and the general partial ranking loss in Eq.~\eqref{eqn:cons_rnk_app}. 
\begin{lem_appendix}{B.1}[Bayes predictor of (partial) ranking loss~\cite{gao2013consistency}]\label{cons_r_bayes}
For all $\mathbf{x}$ and $P(\mathbf{y | \mathbf{x}})$, 
  the set of Bayes predictors w.r.t. the general ranking loss in Eq.~\eqref{eqn:cons_rnk_notpartial_app} is given by 
  \begin{align}
      \mathcal{B}_{L^{0/1}_{gr}}(\mathbf{x}, P(\mathbf{y | \mathbf{x}})) = \{ f: \forall 1 \le p < q \le c, f_p > f_q \textrm{ if } \Delta_{pq}^{+-} >  \Delta_{pq}^{-+}  ; , f_p \neq  f_q \textrm{ if } \Delta_{pq}^{+-} =  \Delta_{pq}^{-+}  ;  f_p < f_q 
      \textrm{ otherwise} \},
  \end{align} 
    and the set of Bayes predictors w.r.t. the general partial ranking loss in Eq.~\eqref{eqn:cons_rnk_app} is given by
  \begin{align}
      \mathcal{B}_{L^{0/1}_{gpr}}(\mathbf{x}, P(\mathbf{y | \mathbf{x}})) = \{ f: \forall 1 \le p < q \le c, f_p > f_q \textrm{ if } \Delta_{pq}^{+-} >  \Delta_{pq}^{-+}  ;  f_p < f_q 
      \textrm{ if }  \Delta_{pq}^{+-} <  \Delta_{pq}^{-+} \}. 
  \end{align} 
\end{lem_appendix}

Similarly to Eq.~\eqref{eqn:cons_pni}, we  define 
\begin{align}
    \phi_p^+  = \sum_{\mathbf{y}: y_p = + 1 }  \beta_\mathbf{y}^+
 P(\mathbf{y}| \mathbf{x})  \textrm{ and } 
    \phi_p^-  = \sum_{\mathbf{y}: y_p = -1 } 	
\beta_\mathbf{y}^- P(\mathbf{y}| \mathbf{x}). 
\end{align}

The following Lemma~\ref{cons_s_bayes} characterizes the set of the Bayes predictors w.r.t. the general  reweighted univariate surrogate loss  in Eq.~\eqref{eqn:our_s_loss} with $\ell(z) = e^{-z}$ or  $\ell(z) = \ln (1 + e^{-z})$.  
\begin{lem_appendix}{B.2}[Bayes predictor of Eq.~\eqref{eqn:our_s_loss} with exponential or logistic loss]\label{cons_s_bayes}
For all $\mathbf{x}$ and $P(\mathbf{y | \mathbf{x}})$, the set of Bayes predictors w.r.t. the general  reweighted univariate surrogate loss in Eq.~\eqref{eqn:our_s_loss} with $\ell(z) = e^{-z}$ or  $\ell(z) = \ln (1 + e^{-z})$ is given by\footnote{Because $\sum_{\mathbf{y}} P(\mathbf{y}|\mathbf{x}) = 1$ for any $\mathbf{x}$ and we assume that the penalties are positive, then $\forall 1 \le j \le c$, $\phi_j^+ + \phi_j^-> 0$.} 
  \begin{align}
      \mathcal{B}_{L_u}^{\ell}(\mathbf{x}, P(\mathbf{y | \mathbf{x}}))  = \{ f: \forall 1 \le j \le c,  f_j = C\ln  \frac{\phi^+_j}{\phi^-_j } \textrm{ if }  \phi_j^+ \phi_j^- > 0;  f_j = + \infty \textrm{ if } \phi_j^- = 0 ;  f_j = -\infty \textrm{ if } \phi_j^+  = 0  \}, 
  \end{align} 
  where $C = \frac{1}{2}$ if  $\ell(z) = e^{-z}$ and $C = 1$ if $\ell(z) = \ln (1 + e^{-z})$.  
\end{lem_appendix}

The following Lemma~\ref{cons_sq_hinge_bayes} and Lemma~\ref{lem_hinge_bayes} 
characterize the set of the Bayes predictors w.r.t. the general  reweighted univariate surrogate loss  in Eq.~\eqref{eqn:our_s_loss} with $\ell(z) = (\max\{0, 1- z\})^2$ and $\ell(z) =  \max\{0, 1- z\} $, respectively.  
\begin{lem_appendix}{B.3}[Bayes predictor of Eq.~\eqref{eqn:our_s_loss} with squared hinge loss]\label{cons_sq_hinge_bayes}
For all $\mathbf{x}$ and $P(\mathbf{y | \mathbf{x}})$, the set of Bayes predictors w.r.t. the general  reweighted univariate surrogate loss in Eq.~\eqref{eqn:our_s_loss} with $\ell(z) = (\max(0, 1- z))^2$ is given by
  \begin{align}
      \mathcal{B}^{\ell}_{L_u}(\mathbf{x}, P(\mathbf{y | \mathbf{x}}))  = \{ f: \forall 1 \le j \le c, f_j = \frac{\phi_j^+ -\phi_j^-}{\phi_j^+ + \phi_j^-} \}.
  \end{align} 
\end{lem_appendix}

\begin{lem_appendix}{B.4}[Bayes predictor of Eq.~\eqref{eqn:our_s_loss} with hinge loss]\label{lem_hinge_bayes}
For all $\mathbf{x}$ and $P(\mathbf{y | \mathbf{x}})$, the set of Bayes predictors w.r.t. the general  reweighted univariate surrogate loss in Eq.~\eqref{eqn:our_s_loss} with $\ell(z) =  \max(0, 1- z) $ is given by
  \begin{align}
      \mathcal{B}^{\ell}_{L_u}(\mathbf{x}, P(\mathbf{y | \mathbf{x}}))  = \{ f: \forall 1 \le j \le c,  f_j =  1  \textrm{ if } \phi_j^+ >\phi_j^- ; f_j = -1  \textrm{ if } \phi_j^+ <\phi_j^- \}.
  \end{align} 
\end{lem_appendix}

The proof of Lemma~\ref{cons_s_bayes}, 
Lemma~\ref{cons_sq_hinge_bayes} and Lemma~\ref{lem_hinge_bayes} are presented in Appendix~\ref{sec:cons_proof_lemma}. 
Combining the Lemma~\ref{cons_cond_app}, Lemma~\ref{cons_r_bayes}, Lemma~\ref{cons_s_bayes} and 
Lemma~\ref{cons_sq_hinge_bayes}, 
we have the following sufficient and necessary condition for the general  reweighted univariate surrogate loss  in Eq.~\eqref{eqn:our_s_loss}  with $\ell(z) = e^{-z}$ or  $\ell(z) = \ln (1 + e^{-z})$ or $\ell(z) = (\max\{0, 1- z\})^2$ to be consistent, as summarized in Proposition~\ref{cons_our_cond}.
\begin{pro_appendix}{B.1}[Sufficient and necessary condition for the consistency of Eq.~\eqref{eqn:our_s_loss} w.r.t. Eq.~\eqref{eqn:cons_rnk_app} with exponential, logistic or squared hinge loss; proof in Appendix~\ref{sec:cons_proof_our_cond}]\label{cons_our_cond}
	The general reweighted univariate surrogate loss in Eq.~\eqref{eqn:our_s_loss} with $\ell(z) = e^{-z}$ or $\ell(z) = \ln (1 + e^{-z})$ or $\ell(z) = (\max\{0, 1- z\})^2$ is consistent w.r.t. the general partial ranking loss in Eq.~\eqref{eqn:cons_rnk_app} if and only if for all $ \mathbf{x}$ and  $P(\mathbf{y} | \mathbf{x})$, we have 
		\begin{align}
		\forall 1 \le p < q \le c, 
 \phi_p^+ \phi_q^-  -  \phi_p^- \phi_q^+  > 0	\textrm{ if }  \Delta_p^+ \Delta_q^-  -  \Delta_p^- \Delta_q^+  > 0 ;	\phi_p^+ \phi_q^-  -  \phi_p^- \phi_q^+  < 0	\textrm{ if }  \Delta_p^+ \Delta_q^-  -  \Delta_p^- \Delta_q^+  < 0.
	\end{align}
\end{pro_appendix}
 
Note that  it takes additional efforts to check the consistency of a new surrogate loss  according to Lemma~\ref{cons_cond_app} or Proposition~\ref{cons_our_cond},  because one has to enumerate all possible conditional distributions. For the general loss in Eq.~\eqref{eqn:our_s_loss}, 
we present more intuitive characterization that only involves the penalties in  Theorem~\ref{thm:cons_log} and Proposition~\ref{cons_pro_hinge},  considering different base losses.

\begin{thm_appendix}{4}[Necessary condition for the consistency of Eq.~\eqref{eqn:our_s_loss} w.r.t. Eq.~\eqref{eqn:cons_rnk_app} with exponential, logistic or squared hinge loss; proof in  Appendix~\ref{sec:cons_proof_thm}]
A general reweighted univariate surrogate loss  in Eq.~\eqref{eqn:our_s_loss} with $\ell(z) = e^{-z}$, $\ell(z) = \ln (1 + e^{-z})$ or $\ell(z) = (\max\{0, 1- z\})^2$ is consistent w.r.t. the general partial ranking loss in Eq.~\eqref{eqn:cons_rnk_app}  only if $\exists \tau  > 0$, $\beta_\mathbf{y}^+ \beta_\mathbf{y}^-  = \tau \alpha_\mathbf{y}^2  $ for all $\mathbf{y}$ such that  $1-c\le \sum_{1\le j \le c}y_j \le c-1$. 
\label{thm:cons_log_app}
\end{thm_appendix}
 
Note that, when $c \le 3$, the penalties of $L_{u_1}$, $L_{u_3}$ and $L_{u_4}$ may coincide with that of $L_{u_2}$ up to a multiplicative constant.
When $c \ge 4$, it is straightforward to construct counter examples that violate the necessary condition in Theorem~\ref{thm:cons_log_app} and obtain the following Corollary~\ref{cons_cor_log_all_app}.
\begin{corollary}[Inconsistency of $L_{u_1}$, $L_{u_3}$ and $L_{u_4}$ w.r.t. Eq.~\eqref{eqn:spe_prl_app} with exponential, logistic or squared hinge loss; proof in  Appendix~\ref{proof_cons_cor}]
If $c \ge 4$, $L_{u_1}$, $L_{u_3}$ and $L_{u_4}$ with  $\ell(z) = e^{-z}$ or $\ell(z) = \ln (1 + e^{-z})$ or $\ell(z) = (\max\{0, 1- z\})^2$
 are inconsistent w.r.t. the partial ranking loss in Eq.~\eqref{eqn:spe_prl_app}. \label{cons_cor_log_all_app}
\end{corollary}

Based on Lemma~\ref{cons_cond_app} and Lemma~\ref{lem_hinge_bayes}, we further show the inconsistency of the general reweighted univariate loss  in Eq.~\eqref{eqn:our_s_loss} w.r.t. the general partial ranking loss in Eq.~\eqref{eqn:cons_rnk_app}  with hinge loss. Note that this includes the inconsistency of $L_{u_1}$, $L_{u_3}$ and $L_{u_4}$ w.r.t. Eq.~\eqref{eqn:spe_prl_app}.

\begin{pro}[Inconsistency of Eq.~\eqref{eqn:our_s_loss} w.r.t. Eq.~\eqref{eqn:cons_rnk_app} with hinge loss; proof in Appendix~\ref{sec:cons_hinge_proof}]
The general reweighted univariate surrogate loss  in Eq.~\eqref{eqn:our_s_loss} with $\ell(z) = \max \{0, 1 - z\}$  are inconsistent w.r.t. the general partial ranking loss in Eq.~\eqref{eqn:cons_rnk_app}, for all positive penalties $\alpha_{\mathbf{y}}$, $\beta^+_{\mathbf{y}}$ and $\beta^-_{\mathbf{y}}$.  \label{cons_pro_hinge_app}
\end{pro}

An immediate conclusion from Corollary~\ref{cons_cor_log_all_app} and Proposition~\ref{cons_pro_hinge_app} is that $L_{u_1}$, $L_{u_3}$ and $L_{u_4}$ are inconsistent w.r.t. the ranking loss in Eq.~\eqref{eqn:spe_rl_app} because $\mathcal{B}_{L_{r}^{0/1}}(\mathbf{x}, P(\mathbf{y | \mathbf{x}})) \subset 	\mathcal{B}_{L_{pr}^{0/1}}(\mathbf{x}, P(\mathbf{y | \mathbf{x}}))$~\cite{gao2013consistency}.
Compared to existing work~\cite{dembczynski2012consistent,gao2013consistency}, although  Theorem~\ref{thm:cons_log_app} and Proposition~\ref{cons_pro_hinge_app} are negative, this paper considers  surrogate losses in a more general reweighted form, i.e. Eq.~\eqref{eqn:our_s_loss}, which may be of independent interest.

\subsection{Proof of Lemma~\ref{cons_s_bayes}, Lemma~\ref{cons_sq_hinge_bayes} and Lemma~\ref{lem_hinge_bayes}}
\label{sec:cons_proof_lemma}

According to Eq.~\eqref{eqn:our_s_loss}, the  conditional risk  for the general  reweighted univariate surrogate loss  in Eq.~\eqref{eqn:our_s_loss} is:
\begin{align}
R (f| \mathbf{x})  & = \sum_{\mathbf{y}}  P(\mathbf{y}| \mathbf{x}) L_{u} (f(\mathbf{x}), \mathbf{y}) \nonumber \\ 
& =  \sum_{\mathbf{y}}  P(\mathbf{y}| \mathbf{x}) \sum_{j=1}^c ([\![ y_j = + 1 ]\!] \beta_\mathbf{y}^+ + [\![ y_j  = - 1 ]\!] \beta_\mathbf{y}^- )  \ell (y_j f_j), \nonumber \\
& =  \sum_{\mathbf{y}} \sum_{j=1}^c ([\![ y_j = + 1 ]\!] \beta_\mathbf{y}^+ + [\![ y_j  = - 1 ]\!] \beta_\mathbf{y}^- )   P(\mathbf{y}| \mathbf{x})  \ell (y_j f_j ), \nonumber \\
& = \sum_{j=1}^c 	 \left [ \sum_{\mathbf{y}: y_j = +1 } 	
 \beta_\mathbf{y}^+  P(\mathbf{y}| \mathbf{x})  \ell (  f_j ) + \sum_{\mathbf{y}: y_j = -1 } \beta_\mathbf{y}^-	
 P(\mathbf{y}| \mathbf{x})  \ell (-f_j ) \right ] \nonumber \\ 
& = \sum_{j=1}^c 	 \left [ \phi_j^{+} \ell (  f_j ) + \phi_j^{-} \ell (- f_j ) \right ]. \label{eqn:our_cond_risk}
\end{align} 

\subsubsection{Proof of Lemma~\ref{cons_s_bayes}.} 
 
\begin{proof}
Because $\sum_{\mathbf{y}} P(\mathbf{y}|\mathbf{x}) = 1$ for any $\mathbf{x}$ and we assume that the penalties are positive, then $\forall 1 \le j \le c$, $\phi_j^+ + \phi_j^-> 0$.
Note that both the exponential loss and logistic loss are  strictly  monotonically  decreasing functions. 

According to Eq.~\eqref{eqn:our_cond_risk},
if $\phi_j^{+} =0$, then $\phi_j^{-} \neq 0$ and $f^*_j(\mathbf{x}) = + \infty$. If  $\phi_j^{-} =0$, then $\phi_j^{+} \neq 0$ and $f^*_j(\mathbf{x}) = - \infty$. We now discuss the case where $\phi_j^+ \phi_j^- > 0$. 

For the exponential loss $\ell(z) =  e^{-z} $, we consider $g(z) = a  e^{-z}  + b  e^{z} $ for $a > 0$ and $b > 0$. It achieves its minima at $z^* = \frac{1}{2} \ln  \frac{a}{b}$. To see this, just take the gradient up to the second order and get 
\begin{align}
    g'(z) =  - a e^{-z} + b e^{-z}, g''(z) = a e^{-z} + b e^{-z}.
\end{align}
Since $\forall z, g''(z) > 0 $. Therefore $g(z)$ is convex. Let $g'(z^*) = 0 \Rightarrow z^* = \frac{1}{2} \ln  \frac{a}{b}$.

For the logistic loss $\ell(z) = \ln (1+e^{-z})$, we consider $g(z) = a \ln  (1 + e^{-z}) + b \ln  (1 + e^{z})$ for $a > 0$ and $b > 0$. It achieves its minima at $z^* = \ln  \frac{a}{b}$ . To see this, just take the gradient up to the second order and get 
\begin{align}
    g'(z) =  \frac{-a e^{-z}}{1+e^{-z}} +\frac{b e^{z}}{1+e^{z}}, g''(z) = \frac{(a+b)e^{z}}{(1+e^z)^2} > 0.
\end{align}
Since $\forall z, g''(z) > 0 $. Therefore $g(z)$ is convex. Let $g'(z^*) = 0 \Rightarrow z^* = \ln  \frac{a}{b}$. Combining all cases together completes the proof.
\end{proof}

\subsubsection{Proof of Lemma~\ref{cons_sq_hinge_bayes}}

\begin{proof}
According to Eq.~\eqref{eqn:our_cond_risk}, the conditional risk of the squared hinge loss $\ell(z) = (\max\{0, 1 - z\})^2$ is 
\begin{align}
    R (f| \mathbf{x})
  = \sum_{j=1}^c 	 \left [ \phi_j^{+} (\max \{0, 1 - f_j \})^2 + \phi_j^{-} (\max \{0, 1 + f_j \})^2 \right ].  
\end{align}

Consider $g(z) = a (\max \{0, 1 - z\})^2 + b (\max \{0, 1 + z \})^2$ for $a \ge 0$, $b \ge 0$ and $a + b > 0$. If $z < - 1$, then $g(z) = a (1 - z)^2 > 4 a$. If $z > 1$, then $g(z) = b (1 + z)^2 > 4 b$. If $-1 \le z \le 1$, then $g(z) = (a + b) z^2 + 2 (b - a) z + (a + b)$, which is convex. The minima is achieved at $z^* = \frac{a - b}{b + a}$, which satisfies $-1 \le z^* \le 1$. The value of $g(z^*)$ is $\frac{4 a b}{a + b} \le \min\{4a, 4b\}$, which means that it is the global minima. Applying this to all $1\le j \le c$ completes the proof.
\end{proof}

\subsubsection{Proof of Lemma~\ref{lem_hinge_bayes}}

\begin{proof}
According to Eq.~\eqref{eqn:our_cond_risk}, the conditional risk of the hinge loss $\ell(z) = \max\{0, 1 - z\}$ is 
\begin{align}
    R (f| \mathbf{x})
  = \sum_{j=1}^c 	 \left [ \phi_j^{+} \max \{0, 1 - f_j \} + \phi_j^{-} \max \{0, 1 + f_j \} \right ].  
\end{align}

Consider $g(z) = a  \max \{0, 1 - z\}   + b  \max \{0, 1 + z \} $ for $a \ge 0$, $b \ge 0$ and $a + b > 0$. If $z < - 1$, then $g(z) = a (1 - z)  > 2 a$. If $z > 1$, then $g(z) = b (1 + z)  > 2 b$. If $-1 \le z \le 1$, then $g(z) = a + b + (b - a)z$. 
If $b > a$, then $z^* = -1$ and $g(z^*) = 2a < 2b$,  which means that it is the global minima.  
If $b < a$, then $z^* = 1$ and $g(z^*) = 2b < 2a$,  which means that it is the global minima. If $b = a$, then whatever $z$ is $g(z) = 2a$. Applying this to all $1\le j \le c$ completes the proof.
\end{proof}

\subsection{Proof of Proposition~\ref{cons_our_cond}} 
\label{sec:cons_proof_our_cond}

\begin{proof}
First, note that $\forall p \neq q, \Delta_p^+ + \Delta_p^- = \Delta_q^+ + \Delta_q^- = \sum_{\mathbf{y}}  \alpha_\mathbf{y}P(\mathbf{y}| \mathbf{x}) > 0$, $\Delta_{pq}^{+-} -  \Delta_{pq}^{-+} = \Delta_p^+  -  \Delta_q^+ $, and 
$$
\Delta_p^+ \Delta_q^-  -  \Delta_p^- \Delta_q^+ =  \Delta_p^+ \left[\sum_{\mathbf{y}}  \alpha_\mathbf{y}P(\mathbf{y}| \mathbf{x}) - \Delta_q^+\right]  -  \left[\sum_{\mathbf{y}}  \alpha_\mathbf{y}P(\mathbf{y}| \mathbf{x}) - \Delta_p^+\right] \Delta_q^+ = \left [ \sum_{\mathbf{y}}  \alpha_\mathbf{y}P(\mathbf{y}| \mathbf{x})\right]  (\Delta_p^+  -  \Delta_q^+ ).
$$ 
Therefore, we have $\forall p < q,$ 
$$\Delta_{pq}^{+-} >  \Delta_{pq}^{-+} \Leftrightarrow
 \Delta_p^+  >  \Delta_q^+ \Leftrightarrow  \Delta_p^+ \Delta_q^-  -  \Delta_p^- \Delta_q^+  > 0,$$ and 
$$\Delta_{pq}^{+-} <  \Delta_{pq}^{-+} \Leftrightarrow  \Delta_p^+  <  \Delta_q^+ 
\Leftrightarrow \Delta_p^+ \Delta_q^-  -  \Delta_p^- \Delta_q^+  < 0 .$$
According to Lemma~\ref{cons_s_bayes}, when  $\ell(z) = e^{-z}$ or  $\ell(z) = \ln (1 + e^{-z})$, $\forall f \in \mathcal{B}^{\ell}_{L_u}(\mathbf{x}, P(\mathbf{y | \mathbf{x}}))$, 
if $  \phi_j^+\phi_j^- > 0$, we have
$  f_j = C\ln  \frac{\phi^+_j}{\phi^-_j},$ 
where $C$ is a constant. Therefore, $\forall 1 \le  p < q \le c$, if $\phi_p^+\phi_p^- > 0$ and  $\phi_q^+\phi_q^- > 0$, then
$f_p > f_q  \Leftrightarrow
\phi_p^+ \phi_q^-  -  \phi_p^- \phi_q^+  < 0$ and  $f_p > f_q \Leftrightarrow
\phi_p^+ \phi_q^-  -  \phi_p^- \phi_q^+  < 0$. 
It is easy to check this also holds if $\phi_p^+ \phi_p^- = 0$ or $\phi_q^+ \phi_q^- = 0$.  Note that we do not need to consider the cases where $\phi_p^+  = \phi_q^+ = 0$ or  $\phi_p^-  = \phi_q^- = 0$ because they imply $\Delta_p^+ \Delta_q^-  -  \Delta_p^- \Delta_q^+  = 0$. Combining with Lemma~\ref{cons_cond_app}, we complete the proof for the  logistic loss and exponential loss.
		
According to Lemma~\ref{cons_sq_hinge_bayes}, when $\ell(z) = (\max\{0, 1- z\})^2$, $\forall f \in \mathcal{B}^{\ell}_{L_u}(\mathbf{x}, P(\mathbf{y | \mathbf{x}}))$, $1 \le j \le c, f_j = \frac{\phi_j^+ -\phi_j^-}{\phi_j^+ + \phi_j^-}$. Therefore, $\forall 1 \le  p < q \le c$,  
$f_p > f_q  \Leftrightarrow  \frac{\phi_p^+ -\phi_p^-}{\phi_p^+ + \phi_p^-} > \frac{\phi_q^+ -\phi_q^-}{\phi_q^+ + \phi_q^-} \Leftrightarrow 
\phi_p^+ \phi_q^-  -  \phi_p^- \phi_q^+  < 0$ and  $f_p > f_q \Leftrightarrow  \frac{\phi_p^+ -\phi_p^-}{\phi_p^+ + \phi_p^-} < \frac{\phi_q^+ -\phi_q^-}{\phi_q^+ + \phi_q^-} \Leftrightarrow 
\phi_p^+ \phi_q^-  -  \phi_p^- \phi_q^+  < 0$. Combining with Lemma~\ref{cons_cond_app}, we complete the proof for the squared hinge loss.  
		
\end{proof}

\subsection{Proof of Theorem~\ref{thm:cons_log_app}}
\label{sec:cons_proof_thm}

\begin{proof}
For convenience, for all $p \neq q$, we define 
\begin{align}
    \phi_{pq}^{rk} = \sum_{\mathbf{y}: y_p = s_r,  y_q = s_k}  ([\![ y_p = + 1 ]\!] \beta_\mathbf{y}^+ + [\![ y_p  = - 1 ]\!] \beta_\mathbf{y}^-)  P(\mathbf{y}| \mathbf{x}), 
\end{align}
where $r, k \in \{+, -\}$ and $s_+ = +1$ and $s_- = -1$. Note that for all $p \neq q$, $\phi_{pq}^{++} = \phi_{qp}^{++} $ and $\phi_{pq}^{--} = \phi_{qp}^{--} $ according to the definition.
For all $1 \le p < q \le c$, we have 
\begin{align}
   \phi_p^+ \phi_q^-  -  \phi_p^- \phi_q^+ & =  (\phi_{pq}^{++} + \phi_{pq}^{+-}) (\phi_{qp}^{-+} +\phi_{qp}^{--})  - (\phi_{pq}^{-+} + \phi_{pq}^{--}) (\phi_{qp}^{++} +\phi_{qp}^{+-}) \nonumber \\
   & = \phi_{pq}^{++}\phi_{qp}^{-+} + \phi_{pq}^{+-} \phi_{qp}^{-+} + \phi_{pq}^{+-} \phi_{qp}^{--} - \phi_{pq}^{-+} \phi_{qp}^{+-} - \phi_{pq}^{-+} \phi_{qp}^{++} -\phi_{pq}^{--} \phi_{qp}^{+-},
   \label{eqn:phis}
\end{align}
and similarly
\begin{align}
   \Delta_p^+ \Delta_q^-  -  \Delta_p^- \Delta_q^+ & =  (\Delta_{pq}^{++} + \Delta_{pq}^{+-}) (\Delta_{qp}^{-+} +\Delta_{qp}^{--})  - (\Delta_{pq}^{-+} + \Delta_{pq}^{--}) (\Delta_{qp}^{++} +\Delta_{qp}^{+-}) \nonumber \\
   & = \Delta_{pq}^{++}\Delta_{qp}^{-+} + \Delta_{pq}^{+-} \Delta_{qp}^{-+} + \Delta_{pq}^{+-} \Delta_{qp}^{--} - \Delta_{pq}^{-+} \Delta_{qp}^{+-} - \Delta_{pq}^{-+} \Delta_{qp}^{++} -\Delta_{pq}^{--} \Delta_{qp}^{+-}.
   \label{eqn:deltas}
\end{align}

For simplicity, we say a $\mathbf{y}$ is \emph{nontrivial} if it satisfies $1-c\le \sum_{1\le j \le c}y_j \le c-1$. 
Assume the consistency holds. We prove that $\exists \tau  > 0$,  $\beta_\mathbf{y}^+ \beta_\mathbf{y}^- = \tau  \alpha_\mathbf{y}^2$ for all nontrivial $\mathbf{y}$. The proof consists of two main steps. 

\textbf{Step 1: } We first prove that, for all $1 \le p < q \le c$, there exists $\tau > 0$, $\beta^+_{\mathbf{y}} \beta^-_{\mathbf{y}} = \tau \alpha_{\mathbf{y}}^2$
for all $\mathbf{y}$ such that $y_p y_q = -1$. According to Proposition~\ref{cons_our_cond}, $\forall \mathbf{x}$ and  $P(\mathbf{y} | \mathbf{x})$, 
		\begin{align}
		\forall p < q, 
 \phi_p^+ \phi_q^-  -  \phi_p^- \phi_q^+  > 0	\textrm{ if }  \Delta_p^+ \Delta_q^-  -  \Delta_p^- \Delta_q^+  > 0 ;	\phi_p^+ \phi_q^-  -  \phi_p^- \phi_q^+  < 0	\textrm{ if }  \Delta_p^+ \Delta_q^-  -  \Delta_p^- \Delta_q^+  < 0 \nonumber.
\end{align}

We simply consider the cases where $P(\mathbf{y}| \mathbf{x}) = 0$
for all $\mathbf{y}$ such that $y_p = y_q.$  According to Eq.~\eqref{eqn:phis} and Eq.~\eqref{eqn:deltas}, we get
\begin{align}
   \phi_p^+ \phi_q^-  -  \phi_p^- \phi_q^+ & =    \phi_{pq}^{+-} \phi_{qp}^{-+}   -\phi_{pq}^{-+} \phi_{qp}^{+-}  \nonumber \\
  & =  \sum_{\substack{\mathbf{y}: y_p = +1, y_q = -1 \\ \mathbf{y}': y'_p = +1, y'_q = -1 }} \beta_{\mathbf{y}}^+ \beta_{\mathbf{y}'}^- P(\mathbf{y}| \mathbf{x}) P(\mathbf{y}'| \mathbf{x})
  - \sum_{\substack{\mathbf{y}: y_p = -1, y_q = +1 \\ \mathbf{y}': y'_p = -1, y'_q = +1 }} \beta_{\mathbf{y}}^+ \beta_{\mathbf{y}'}^- P(\mathbf{y}| \mathbf{x}) P(\mathbf{y}'| \mathbf{x}),
   \label{eqn:phis_2}
\end{align}
and
\begin{align}
   \Delta_p^+ \Delta_q^-  -  \Delta_p^- \Delta_q^+ & =    \Delta_{pq}^{+-} \Delta_{qp}^{-+}   -\Delta_{pq}^{-+} \Delta_{qp}^{+-} \nonumber \\
  & =  \sum_{\substack{\mathbf{y}: y_p = +1, y_q = -1 \\ \mathbf{y}': y'_p = +1, y'_q = -1 }} \alpha_{\mathbf{y}} \alpha_{\mathbf{y}'} P(\mathbf{y}| \mathbf{x}) P(\mathbf{y}'| \mathbf{x})
  - \sum_{\substack{\mathbf{y}: y_p = -1, y_q = +1 \\ \mathbf{y}': y'_p = -1, y'_q = +1 }} \alpha_{\mathbf{y}} \alpha_{\mathbf{y}'} P(\mathbf{y}| \mathbf{x}) P(\mathbf{y}'| \mathbf{x}).
   \label{eqn:deltas_2}
\end{align}

We proceed by contradiction and consider two cases. Recall that we assume $\alpha_{\mathbf{y}} > 0$ and $  \beta^+_{\mathbf{y} }\beta^-_{\mathbf{y} }  >  0$ for all  nontrivial  $\mathbf{y}$. Suppose that there exists $\tau_3 > 0,  \tau_4 > 0,  \tau_3 \neq \tau_4$, $ \beta^+_{\mathbf{y}}\beta^-_{\mathbf{y}}
= \tau_3 \alpha_{\mathbf{y}}^2       \neq \tau_4 \alpha_{\mathbf{y}}^2 $
and 
$  \beta^+_{\mathbf{y}'}\beta^-_{\mathbf{y}'} = \tau_4  \alpha_{\mathbf{y}'}^2  \neq \tau_3 \alpha_{\mathbf{y}'}^2 $ for some $\mathbf{y} \neq \mathbf{y}'$ with $y_p y_q = -1$ and $y'_p y'_q = -1$. 



\textbf{Case 1.1:}  
$y_p \neq y'_p$. Without loss of generality, let $y_p = +1$  and $y'_p = -1$. Accordingly, we get  $y_q = -1$ and $y'_q = +1$. Let  $P(\mathbf{y}|\mathbf{x}) = \frac{\sqrt{\beta^+_{\mathbf{y}'}\beta^-_{\mathbf{y}'}}}{\sqrt{\beta^+_{\mathbf{y}}\beta^-_{\mathbf{y} } } + \sqrt{\beta^+_{\mathbf{y}'}\beta^-_{\mathbf{y}'}}}$ and $ P(\mathbf{y}'|\mathbf{x}) = \frac{\sqrt{\beta^+_{\mathbf{y}}\beta^-_{\mathbf{y} } } }{\sqrt{\beta^+_{\mathbf{y}}\beta^-_{\mathbf{y} } } + \sqrt{\beta^+_{\mathbf{y}'}\beta^-_{\mathbf{y}'}}}$. Note that $P(\mathbf{y}|\mathbf{x})  + P(\mathbf{y}'|\mathbf{x}) = 1$. According to Eq.~\eqref{eqn:deltas_2}, we have
$$ 
\Delta_p^+ \Delta_q^-  -  \Delta_p^- \Delta_q^+ = \alpha_{\mathbf{y}}^2  P(\mathbf{y}| \mathbf{x}) ^2 -  \alpha_{\mathbf{y}'}^2  P(\mathbf{y}'| \mathbf{x}) ^2
= \frac{\beta^+_{\mathbf{y} }\beta^-_{\mathbf{y} }\beta^+_{\mathbf{y}'}\beta^-_{\mathbf{y}'} }{(\sqrt{\beta^+_{\mathbf{y}}\beta^-_{\mathbf{y} } } + \sqrt{\beta^+_{\mathbf{y}'}\beta^-_{\mathbf{y}'}})^2} (\frac{1}{\tau_4} - \frac{1}{\tau_3}) \neq 0,
$$
but according to Eq.~\eqref{eqn:phis_2}, we have
$$
\phi_p^+ \phi_q^-  -  \phi_p^- \phi_q^+ = 
\beta_{\mathbf{y}}^+ \beta_{\mathbf{y} }^- P(\mathbf{y}| \mathbf{x}) ^2 - 
\beta_{\mathbf{y}'}^+ \beta_{\mathbf{y}'}^- P(\mathbf{y}'| \mathbf{x}) ^2
= 0,
$$ 
which is a contrary to Proposition~\ref{cons_our_cond}.

\textbf{Case 1.2:} 
$y_p = y'_p$. Without loss of generality, let $y_p = y'_p = -1$, then $y_q = y'_q = + 1$. Consider $y''_p$ such that $y''_p =  +1,  y''_q = -1$. 
Then according to the Case 1.1, there exists $\tau$, such that $\tau = \frac{\beta^+_{\mathbf{y}}\beta^-_{\mathbf{y}}}{\alpha_{\mathbf{y}}^2} = \frac{\beta^+_{\mathbf{y}''}\beta^-_{\mathbf{y}''}}{\alpha_{\mathbf{y}''}^2} = 
\frac{\beta^+_{\mathbf{y}'}\beta^-_{\mathbf{y}'}}{\alpha_{\mathbf{y}'}^2}$, which is a contrary. 

Combining the Case 1.1 and Case 1.2 together, for all $1 \le p < q \le c$, there exists $\tau > 0$,  $\beta^+_{\mathbf{y}} \beta^-_{\mathbf{y}}  = \tau \alpha_{\mathbf{y}}^2$
for all $\mathbf{y}$ such that $y_p y_q = -1$.
 
\textbf{Step 2:} Note that the values of $ \tau$ in Step 1 may depend on $p$ and $q$. Now
we prove that there exists a  universal $ \tau$  for all $1 \le p < q \le c$. For any nontrivial $\mathbf{y} \neq  \mathbf{y}'$, we can find $1 \le p < q \le c$ and $1 \le p'< q' \le c$ such that $y_p y_q = -1$ and $y'_{p'} y'_{q'} = -1$. 
We consider four cases.  
  
\textbf{Case 2.1:} Two pair of indices match, namely, $p=p'$, $q = q'$. We have proven that $\frac{ \beta_{\mathbf{y}}^-
  \beta_{\mathbf{y}}^+ }{  
\alpha_{\mathbf{y}}^2} = \frac{  \beta^-_{\mathbf{y}'} \beta^+_{\mathbf{y}'}}
 {\alpha_{\mathbf{y}'}^2}$ in Step 1.

\textbf{Case 2.2:} No index matches for $c \ge 4$, namely, $p \neq p'$, $q \neq q'$, $p \neq q'$, $p'\neq q$. We can construct $\mathbf{y}''$ such that $y''_p = y_p$, $y''_q = y_q$, $y''_{p'} = y'_{p'}$, $y''_{q'} = y'_{q'}$ and get $\frac{ \beta_{\mathbf{y}}^-\beta_{\mathbf{y}}^+
 }{  
\alpha_{\mathbf{y}}^2} = \frac{  \beta^-_{\mathbf{y}''} \beta^+_{\mathbf{y}''}
 }{\alpha_{\mathbf{y}''}^2}  = \frac{  \beta^-_{\mathbf{y}'}\beta^+_{\mathbf{y}'}
 }{\alpha_{\mathbf{y}'}^2} $ according to Step 1.

\textbf{Case 2.3:} Only one pair of indices match and the corresponding labels are  the same for $c \ge 3$. Without loss of generality, suppose $p=1$, $q = p'=2$, $q' = 3$ and $ y_2 = +1,  y_2' = +1$. It implies that $y_1 = -1$ and $y_3' = - 1$. Suppose $y_3 = -1$, then $ y_2 y_3 =y_2'  y_3' = -1.$  Suppose $y_3 = +1$, no matter which label $y'_1$ is, either $ y_1 y_2 =y_1'  y_2' = -1$ or $ y_1 y_3 =y_1'  y_3' = -1.$ We get  $\frac{ \beta_{\mathbf{y}}^-\beta_{\mathbf{y}}^+
 }{  
\alpha_{\mathbf{y}}^2}  = \frac{  \beta^-_{\mathbf{y}'}\beta^+_{\mathbf{y}'}
 }{\alpha_{\mathbf{y}'}^2} $ according to Step 1.

\textbf{Case 2.4:} Only one pair of indices match and the corresponding labels are not the same for $c \ge 3$. Without loss of generality, suppose $p=1$, $q = p'=2$, and $q' = 3$ and $ y_2 = +1,  y_2' = -1$. We have $y_1 = -1, y_3' = + 1$. Similarly to Case 3, no matter which labels $ y_3$ and $ y'_1 $ are, we have either $ y_1 y_2 =y_1'  y_2' = -1$ or 
$ y_1 y_3 =y_1'  y_3' = -1$ or 
$ y_2 y_3 =y_2'  y_3' = -1$, and get $\frac{ \beta_{\mathbf{y}}^-\beta_{\mathbf{y}}^+
 }{  
\alpha_{\mathbf{y}}^2} = \frac{  \beta^-_{\mathbf{y}'}\beta^+_{\mathbf{y}'}
 }{\alpha_{\mathbf{y}'}^2} $ according to Step 1.

Combining Case 2.1, Case 2.2, Case 2.3 and Case 2.4 together, we obtain that for all nontrivial $\mathbf{y} \neq \mathbf{y}'$,  $\frac{ \beta_{\mathbf{y}}^-\beta_{\mathbf{y}}^+
 }{  
\alpha_{\mathbf{y}}^2}    = \frac{  \beta^-_{\mathbf{y}'}\beta^+_{\mathbf{y}'}
 }{\alpha_{\mathbf{y}'}^2} $.
\end{proof}

\subsection{Proof of Corollary~\ref{cons_cor_log_all_app}}
\label{proof_cons_cor}

\begin{proof}
We consider a multi-label classification problem with $c \ge 4$ labels. Let $\mathbf{y}$ satisfy  $y_1 = +1$, and $y_j = -1$ for all $2 \le j \le c$, and $\mathbf{y}'$ satisfy  $y'_1 = y'_2 = +1$, and $y'_j = -1$ for all $ 3 \le j \le c$. 
According to the definition of the partial ranking loss  in Eq.~(\ref{eqn:spe_prl_app}), we have $\alpha_{\mathbf{y}} = \frac{1}{c-1}$ and  $\alpha_{\mathbf{y}'} = \frac{1}{2(c-2)}$. 

In $L_{u_1}$, according to the definition, we have $\beta_\mathbf{  y}^+ = \beta_\mathbf{  y'}^+ =\beta_\mathbf{\hat y}^- = \beta_\mathbf{  y'}^- =  \frac{1}{c}$. It is easy to check that 
$\frac{\beta_\mathbf{y}^+\beta_\mathbf{y}^- }{\alpha_{\mathbf{y}}^2} = \frac{(c-1)^2}{c^2}  \neq  \frac{4(c-2)^2}{c^2} = \frac{\beta_\mathbf{y'}^+\beta_{\mathbf{y}'}^- }{\alpha_{\mathbf{y}'}^2}$ for all $c\ge4$. 

In $L_{u_3}$, according to the definition, we have
$\beta_\mathbf{y}^+ = 1$, $\beta_\mathbf{y'}^+ = \frac{1}{2}$,
$\beta_\mathbf{y}^- = \frac{1}{c-1}$, and $\beta_\mathbf{y'}^- = \frac{1}{c-2}$. It is easy to check  that 
$\frac{\beta_\mathbf{y}^+\beta_\mathbf{y}^- }{\alpha_{\mathbf{y}}^2} =  c-1  \neq  2(c-2)  = \frac{\beta_\mathbf{y'}^+\beta_{\mathbf{y}'}^- }{\alpha_{\mathbf{y}'}^2} $ for all $c\ge4$. 
 
In $L_{u_4}$, according to the definition, we have   
$\beta_\mathbf{y}^+ = 1$, $\beta_\mathbf{y'}^+ = \frac{1}{2}$,
$\beta_\mathbf{y}^- = 1$, and $\beta_\mathbf{y'}^- = \frac{1}{2}$   , for all $c\ge4$. It is easy to check that
$\frac{\beta_\mathbf{y}^+\beta_\mathbf{y}^- }{\alpha_{\mathbf{y}}^2} = (c-1)^2  \neq  (c-2)^2 = \frac{\beta_\mathbf{y'}^+\beta_{\mathbf{y}'}^- }{\alpha_{\mathbf{y}'}^2}  $.

According to Theorem~\ref{thm:cons_log_app} and Proposition~\ref{cons_our_cond}, the above surrogate losses are not consistent w.r.t. the partial ranking loss in Eq.~(\ref{eqn:spe_prl_app}).
\end{proof}

\subsection{Proof of Proposition~\ref{cons_pro_hinge_app}}
\label{sec:cons_hinge_proof}

\begin{proof}

We consider a multi-label classification problem with $c = 2$ labels. Let 
\begin{align*}
    \mathbf{y}_1 = (+1, + 1),  \mathbf{y}_2 = (+1, -1), 
     \mathbf{y}_3 = (+1, - 1),  \mathbf{y}_4 = (-1, -1). 
\end{align*}

Given a data point $\mathbf{x}$, let $0 < \epsilon < \frac{\beta_{ \mathbf{y}_1}^+}{\beta_{ \mathbf{y}_1}^+ + \max \{\beta_{ \mathbf{y}_2}^-, \beta_{ \mathbf{y}_3}^- \}}$. Consider a conditional distribution such that $P( \mathbf{y}_2| \mathbf{x}) P( \mathbf{y}_3|\mathbf{x}) > 0$, $\alpha_{ \mathbf{y}_2} P( \mathbf{y}_2| \mathbf{x}) \neq  \alpha_{ \mathbf{y}_3} P( \mathbf{y}_3|\mathbf{x})$, $P( \mathbf{y}_2| \mathbf{x}) + P( \mathbf{y}_3|\mathbf{x}) = \epsilon$, $P( \mathbf{y}_1| \mathbf{x}) = 1 - \epsilon$ and $P( \mathbf{y}_4| \mathbf{x}) = 0$. On one hand, we get 
\begin{align}
    \Delta_1^+ - \Delta_2^+ = \alpha_{ \mathbf{y}_2} P( \mathbf{y}_2| \mathbf{x}) -  \alpha_{ \mathbf{y}_3} P( \mathbf{y}_3|\mathbf{x}) \neq 0,
\end{align}
which implies $f_1 \neq f_2$ for any $f \in \mathcal{B}_{L^{0/1}_{gpr}}(\mathbf{x}, P(\mathbf{y | \mathbf{x}}))$ according to Lemma~\ref{cons_r_bayes}. On the other hand, we get 
\begin{align}
    \phi_1^+ - \phi_1^- & = \beta_{ \mathbf{y}_1}^+ P( \mathbf{y}_1| \mathbf{x}) +  \beta_{ \mathbf{y}_2}^+ P( \mathbf{y}_2| \mathbf{x}) - \beta_{ \mathbf{y}_3}^- P( \mathbf{y}_3| \mathbf{x}) - \beta_{ \mathbf{y}_4}^- P( \mathbf{y}_4| \mathbf{x}) \nonumber \\ 
    & >  \beta_{ \mathbf{y}_1}^+ P( \mathbf{y}_1| \mathbf{x})   - \beta_{ \mathbf{y}_3}^- P( \mathbf{y}_3| \mathbf{x}) \nonumber \\
        & > \beta_{ \mathbf{y}_1}^+ (1 - \epsilon)
    - \beta_{ \mathbf{y}_3}^- \epsilon \nonumber \\
    & =\beta_{ \mathbf{y}_1}^+ (1 -
     \frac{\beta_{ \mathbf{y}_1}^+}{\beta_{ \mathbf{y}_1}^+ + \max \{\beta_{ \mathbf{y}_2}^-, \beta_{ \mathbf{y}_3}^- \}}
    )   - \beta_{ \mathbf{y}_3}^- \frac{\beta_{ \mathbf{y}_1}^+}{\beta_{ \mathbf{y}_1}^+ +  \max \{\beta_{ \mathbf{y}_3}^-, \beta_{ \mathbf{y}_3}^- \}}  
    \nonumber \\
    & =   \frac{\beta_{ \mathbf{y}_1}^+  (\max \{  \beta_{ \mathbf{y}_2}^- ,\beta_{ \mathbf{y}_3}^- \} - \beta_{ \mathbf{y}_3}^-) }{\beta_{ \mathbf{y}_1}^+ + \max \{\beta_{ \mathbf{y}_2}^-, \beta_{ \mathbf{y}_3}^- \}} \nonumber \\ 
    & \ge 0,
\end{align}
which means that $\forall f  \in  \mathcal{B}^{\ell}_{L_u}(\mathbf{x}, P(\mathbf{y | \mathbf{x}}))$, $f_1 = -1$ according to Lemma~\ref{lem_hinge_bayes}. Similarly, $\forall f  \in  \mathcal{B}_{L_u} (\mathbf{x}, P(\mathbf{y | \mathbf{x}}))$, $f_2 = -1 = f_1$. Therefore, $\mathcal{B}^{\ell}_{L_u}(\mathbf{x}, P(\mathbf{y | \mathbf{x}})) \not \subset \mathcal{B}_{L^{0/1}}(\mathbf{x}, P(\mathbf{y | \mathbf{x}})),$ which completes the proof combining with Lemma~\ref{cons_cond_app}.
\end{proof}

\section{Additional Experimental Results}

The complete experimental results (with standard deviations) are summarized in Table \ref{tab:benchmark_results_full}.
\begin{table}[t]
\renewcommand\tabcolsep{4pt}
\scriptsize
\caption{Ranking loss ($\textrm{mean} \pm \textrm{std}$) of all five algorithms on benchmark datasets. On each dataset, the top two algorithms are highlighted in bold and the top one is labeled with $^{\dagger}$. }
\label{tab:benchmark_results_full}
\vskip 0.15in
\begin{center}
\begin{small}
\begin{tabular}{lccccc}
\toprule
Dataset & $\mathcal{A}^{pa}$ & $\mathcal{A}^{u_1}$ & $\mathcal{A}^{u_2}$ & $\mathcal{A}^{u_3}$ & $\mathcal{A}^{u_4}$ \\
\midrule
emotions & $\bf 0.1511 \pm 0.0175^{\dagger}$ & $0.1538 \pm 0.0219$ & $0.1587 \pm 0.0198$ & $\bf0.1530 \pm 0.0193$ & $0.1616 \pm 0.0202$ \\
image & $\bf 0.1625 \pm 0.0089^{\dagger}$ & $0.1642 \pm 0.0132$ & $0.1653 \pm 0.0153$ & $\bf0.1645 \pm 0.0159$ & $0.1678 \pm 0.0056$ \\
scene & $\bf 0.0696 \pm 0.0031^{\dagger}$ & $0.0809 \pm 0.0083$ & $0.0821 \pm 0.0029$ & $\bf0.0768 \pm 0.0082$ & $0.0806 \pm 0.0025$ \\
yeast & $\bf 0.1766 \pm 0.0078^{\dagger}$ & $0.1768 \pm 0.0093$ & $0.1785 \pm 0.0090$ & $\bf0.1767 \pm 0.0086$ & $0.1816 \pm 0.0084$ \\
enron & $\bf 0.0682 \pm 0.0030^{\dagger}$ & $0.0724 \pm 0.0022$ & $\bf0.0696 \pm 0.0011$ & $0.0698 \pm 0.0027$ & $0.0715 \pm 0.0038$ \\
rcv1-subset1 & $\bf 0.0361 \pm 0.0015^{\dagger}$ & $0.0418 \pm 0.0005$ & $0.0392 \pm 0.0003$ & $\bf0.0368 \pm 0.0003$ & $0.0391 \pm 0.0005$ \\
bibtex & $\bf0.0516 \pm 0.0014$ & $0.0545 \pm 0.0018$ & $0.0551 \pm 0.0024$ & $\bf 0.0401 \pm 0.0694^{\dagger}$ & $0.0538 \pm 0.0020$ \\
corel5k & $\bf0.1081 \pm 0.0021$ & $0.1091 \pm 0.0004$ & $0.1099 \pm 0.0016$ & $\bf 0.1063 \pm 0.0019^{\dagger}$ & $0.1096 \pm 0.0010$ \\
mediamill & $\bf0.0395 \pm 0.0011$ & $0.0402 \pm 0.0005$ & $0.0412 \pm 0.0001$ & $\bf 0.0389 \pm 0.0006^{\dagger}$ & $0.0405 \pm 0.0010$ \\
delicious & - & $\bf0.0960 \pm 0.0010$ & $0.0974 \pm 0.0007$ & $\bf 0.0946 \pm 0.0002^{\dagger}$ & $0.0978 \pm 0.0008$ \\
\bottomrule
\end{tabular}
\end{small}
\end{center}
\vskip -0.1in
\end{table}

\begin{figure}[th!]
\vskip 0.2in
\begin{center}
\centerline{\includegraphics[scale=0.5]{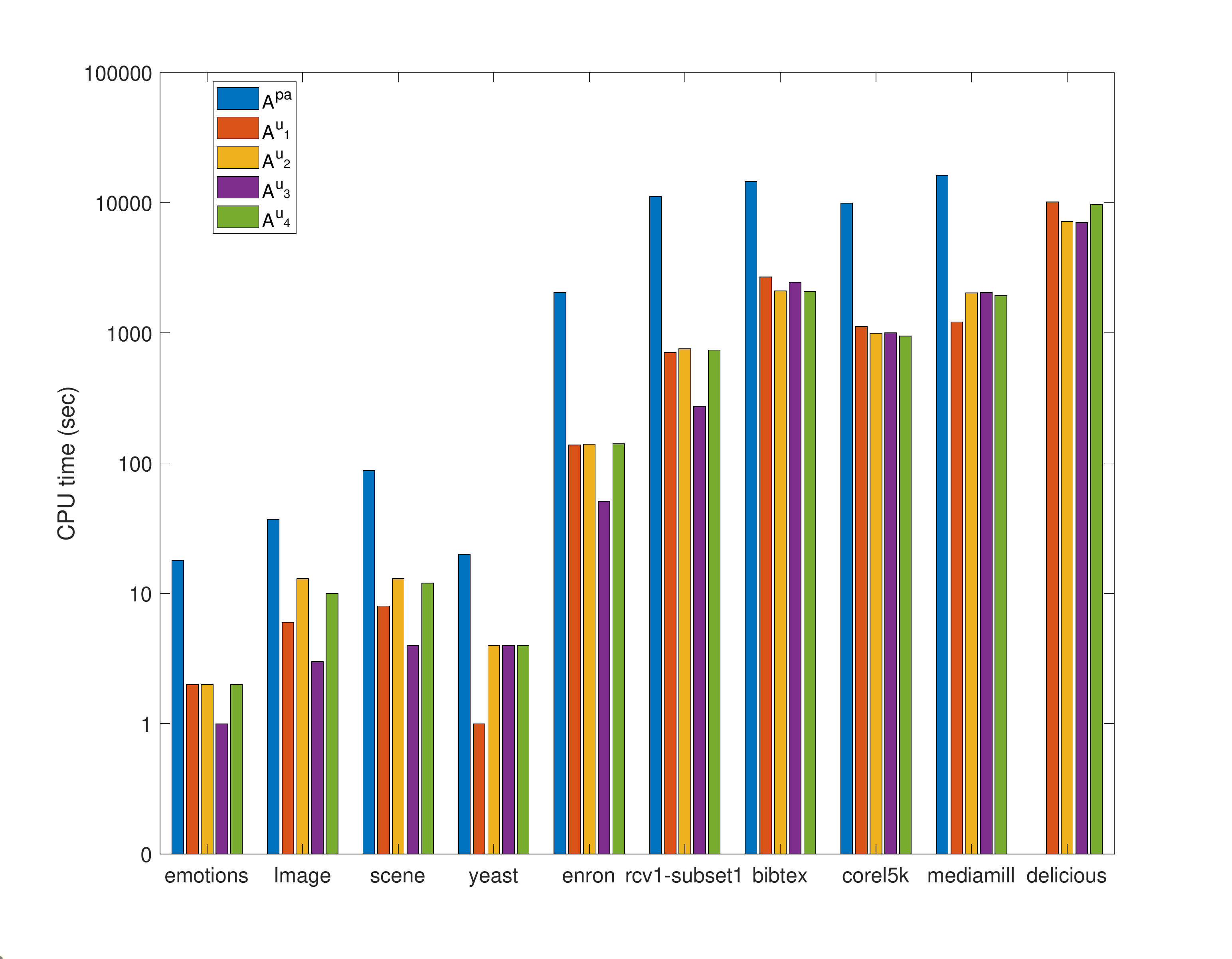}}
\caption{Computational costs of all five algorithms on benchmark datasets.}
\label{fig:cpu_time}
\end{center}
\vskip -0.2in
\end{figure}

Besides, the computational costs of all five algorithms on benchmark datasets are shown in Figure~\ref{fig:cpu_time}. From Figure~\ref{fig:cpu_time}, we can observe that $\mathcal{A}^{pa}$ with the pairwise loss is much slower than the other four algorithms with the univariate loss, especially when the label space is large. Note that the CPU time is plotted in the log scale in Figure~\ref{fig:cpu_time}.

\end{document}